\documentclass{article}
\usepackage{iclr2027_conference,times}

  \iclrfinalcopy
  \usepackage{etoolbox}
  \makeatletter
  \patchcmd{\@maketitle}
    {Published as a conference paper at ICLR 2027}
    {Preprint}
    {}
    {\PackageError{opd-preprint}{Could not set the preprint header}{}}
  \makeatother

\usepackage{amsmath,amsfonts,bm}

\def\eqref#1{equation~\ref{#1}}

\def\1{\bm{1}}

\DeclareMathAlphabet{\mathsfit}{\encodingdefault}{\sfdefault}{m}{sl}
\SetMathAlphabet{\mathsfit}{bold}{\encodingdefault}{\sfdefault}{bx}{n}

\usepackage{url}
\usepackage{xurl}
\usepackage{graphicx}
\usepackage{booktabs}
\usepackage{longtable}
\usepackage{amsthm}
\usepackage{xcolor}
\usepackage{colortbl}
\usepackage{multirow}
\usepackage{placeins}
\usepackage{float}
\usepackage{flafter}
\floatstyle{ruled}
\newfloat{algorithm}{tbp}{loa}
\floatname{algorithm}{Algorithm}
\usepackage{hyperref}
\hypersetup{
  hidelinks,
  pdftitle={Teaching a Moving Student: Rethinking the Curriculum of On-Policy Distillation}
}

  \hypersetup{
    pdfauthor={Lingxiang Hu, Tianle Xia, Yiding Sun, Ming Xu, Linfang Shang, Lan Xu, Ning Zheng, Wei Xu, Jie Jiang}
  }

\theoremstyle{definition}

\theoremstyle{remark}

\newcommand{\DKL}{D_{\mathrm{KL}}}

\newcommand{\Ctraj}{C}

\newcommand{\ropd}{R-OPD}

\title{Teaching a Moving Student: Rethinking\\
the Curriculum of On-Policy Distillation}

  \author{%
  \begin{tabular}[t]{@{}l@{\hspace{1.8em}}l@{\hspace{1.8em}}l@{\hspace{1.8em}}l@{\hspace{1.8em}}l@{}}
    \textbf{Lingxiang Hu\thanks{First author.}} &
    \textbf{Tianle Xia\footnotemark[1]} &
    \textbf{Yiding Sun} &
    \textbf{Ming Xu} &
    \textbf{Linfang Shang}\\[3pt]
    \textbf{Lan Xu\thanks{Corresponding author.}} &
    \textbf{Ning Zheng} &
    \textbf{Wei Xu} &
    \textbf{Jie Jiang} & \\[5pt]
  \end{tabular}\\
  Tencent\\[2pt]
  {\small\texttt{\{lingxianghu,tianlexia,emanuelsun,flemingxu,faelynshang\}@tencent.com}}\\
  {\small\texttt{\{lanxu,yodazheng,davidxu,zeus\}@tencent.com}}}
  \date{}

\begin{document}

\maketitle
\raggedbottom

\begin{abstract}
In on-policy distillation (OPD), the student not only learns from the
teacher but also determines which states receive future supervision.
As the student evolves, training moves to new response prefixes while
teacher--student disagreement can remain on earlier ones.
Under matched trajectory and optimization budgets, neither more training
queries nor more frequent rollout resampling is uniformly beneficial.
Fresh current-policy rollouts outperform initial-policy rollouts at shorter
response budgets, but the ranking reverses at longer budgets.
Fixed-prefix measurements show that earlier prefixes become less likely
under the student while teacher--student disagreement on them persists.
Replaying initial-policy rollouts after current-policy training improves
accuracy, whereas replaying fixed recent rollouts does not reproduce the
gain.
We therefore propose \ropd{}, a gradient-triggered curriculum that lets the
student learn on current-policy states before adaptively introducing replay,
without prescribing a fixed transition iteration. Once mean gradient changes
fall within minibatch-level variation, training switches to initial-policy
rollouts from the next iteration through the remaining training budget.
Across eight mathematics benchmarks, \ropd{}
improves average accuracy over continued current-policy sampling by 2.25
and 4.44 percentage points at 16K and 32K for a 0.6B student across three
training runs, and by 2.81 and 6.15 points for a 1.7B student.
With a 30B-A3B teacher, \ropd{} also improves 8B accuracy by 4.05 points at 32K.
Fixed two-stage replay also improves accuracy. At 32K, \ropd{} exceeds
a fixed schedule of 40 current-policy updates followed by 20 replay updates
by 1.39/1.66/1.85 points for
0.6B/1.7B/8B, averaged over three training-data orders.
At the same 32K generation cap, \ropd{} also produces longer responses,
suggesting that well-timed revisits help the student use more of its
reasoning capacity.
\end{abstract}

\section{Introduction}
\label{sec:intro}

In on-policy distillation (OPD), the student samples a response and the
teacher provides token-level supervision along that response
\citep{gu2024minillm,agarwal2024gkd}. Both models predict the next token
conditioned on the query and the response generated so far, or
\emph{response prefix}. Together, the query and prefix define the
\emph{state} at which supervision is applied.

This coupling creates a distinctive property of OPD: the student not only
learns from the teacher, but also determines which states will receive
future supervision. As the student changes, newly sampled responses move
training toward new prefixes. Yet moving away from an earlier prefix does
not imply that the student has already matched the teacher there.
The student can therefore move on before supervision on earlier states is
exhausted.

This raises our central question:
\emph{how should OPD organize supervision over a state distribution that
changes with the student itself?}

\begin{figure}[t]
  \centering
  \includegraphics[width=\textwidth]{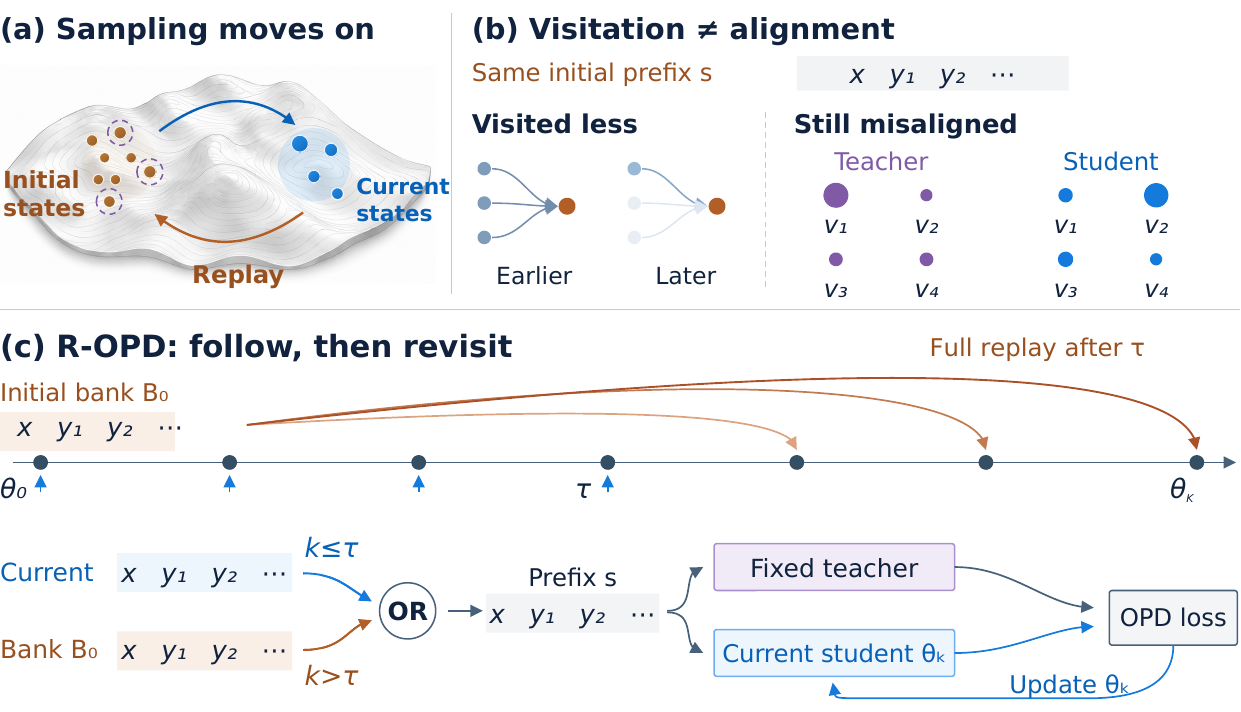}
  \caption{\textbf{\ropd{} follows the student, then revisits earlier states.}
  (a) Current-policy sampling follows the evolving student; replay returns
  to initial-policy states.
  (b) Earlier prefixes become less likely while teacher--student
  predictions remain misaligned.
  (c) At the gradient-triggered iteration $\tau$, \ropd{} finishes the
  current-policy stage; every later iteration uses the initial-policy bank.}
  \label{fig:overview}
\end{figure}

A natural intuition is to expose the student to more queries and to refresh
rollouts as frequently as possible. Our first experiments challenge this
intuition. We jointly vary the number of training queries and the frequency
of rollout resampling while holding trajectory and optimization-step
budgets fixed. With rollouts refreshed every several optimization steps,
only eight queries approach the accuracy obtained with the full dataset
(Table~\ref{tab:surface}). Refreshing rollouts at every step does not
consistently improve accuracy. More strikingly, increasing the number of
queries helps under per-step resampling but hurts when all rollouts are
generated before training.

These results suggest that data breadth alone does not determine the value
of OPD supervision; the policy that generates the supervised states also
matters. We therefore compare rollouts sampled from the evolving student
(\emph{current-policy rollouts}) with a fixed set sampled before OPD
training (\emph{initial-policy rollouts}). At shorter response-length
budgets, current-policy rollouts yield higher accuracy under two different
teachers. At 32K, however, the ordering reverses: training on initial-policy
rollouts becomes more accurate and produces longer responses
(Figure~\ref{fig:budgetcrossover}a). Freshness is therefore not uniformly
advantageous.

Why can earlier rollouts remain useful after the student has moved on?
We separate two quantities that standard on-policy sampling conflates:
whether an earlier prefix is still likely to be visited, and whether the
student has already learned the teacher's prediction at that prefix.
Using prefixes saved from initial-policy rollouts
(Section~\ref{sec:initial-states}), we find that most become less likely
under the trained student. At the same time, the student's predictions on
those prefixes move closer to the teacher's but remain imperfectly aligned.
Earlier states are thus disappearing from the training distribution before
their teacher supervision has been fully absorbed.

To test whether replay improves the trained student, we continue from
the same on-policy-trained checkpoint either with fresh rollouts or
with an \emph{initial-policy rollout bank}, a stored set of responses
generated before OPD training. Replaying the initial-policy bank further
reduces teacher--student disagreement and improves accuracy at both 16K
and 32K (Table~\ref{tab:state-source-core}). Freezing a newly generated
bank from the trained student does not reproduce this gain, showing that
the benefit is not simply a consequence of stopping resampling.

The timing of replay matters as well. With the same total number of
optimization steps allocated to current-policy and initial-policy
supervision, training first on current-policy rollouts and replaying the
initial-policy bank afterward outperforms the reverse order at 16K and 32K
(Table~\ref{tab:state-order-core}). The two sources are therefore not
interchangeable: current-policy supervision is most useful first, while
earlier states become more useful to revisit later.

This temporal asymmetry motivates \ropd{}, an adaptive replay curriculum
(Figure~\ref{fig:overview}). The aim is to let the student learn from teacher
supervision on current-policy states before revisiting initial states,
without manually choosing a transition iteration for each setting.
We monitor mean gradient changes relative to minibatch variation as a
heuristic indication that current-policy learning is stabilizing.
After two consecutive qualifying comparisons at iteration $\tau$, every
iteration from $\tau+1$ onward uses the initial-policy bank.

Across eight mathematics benchmarks, \ropd{} improves average accuracy
over continued on-policy sampling by 2.25 and 4.44 percentage points at
16K and 32K, respectively, averaged over three 0.6B training runs.
For a 1.7B student, the mean gains are 2.81 and 6.15 points.
With a 30B-A3B teacher, \ropd{} also improves mean 8B accuracy by 4.05 points at 32K.
Under the same evaluation settings, these gains accompany longer responses,
consistent with well-timed replay helping the student express more of its
reasoning capacity within the available generation budget.
A fixed schedule of 40 current-policy updates followed by 20 replay
updates (denoted $R_{40}F_{20}$) also improves accuracy. At 32K,
\ropd{} exceeds this schedule by 1.39/1.66/1.85 points for
0.6B/1.7B/8B across three training-data orders.

\begin{samepage}
Our contributions are:

\begin{itemize}

  \item \textbf{Freshness and data breadth interact in OPD.}
  Increasing the number of queries is beneficial under some rollout
  schedules and harmful under others, and the accuracy ordering between
  current-policy and initial-policy rollouts reverses as the response-length
  budget increases.

  \item \textbf{Earlier student states can remain useful after the policy
  moves on.}
  Their generation probability decreases, yet teacher--student disagreement
  persists. Replaying these states after current-policy training reduces
  this disagreement and improves accuracy, while reversing the order gives
  weaker results at 16K and 32K.

  \item \textbf{We turn this temporal asymmetry into a replay curriculum.}
  \ropd{} uses gradient drift to choose when to introduce initial-policy
  replay and improves average accuracy over continued current-policy sampling
  at 16K and 32K for all three student sizes.

\end{itemize}
\end{samepage}

\section{Setup: States, Supervision, and Training}
\label{sec:setup}

\paragraph{States.}
For a query $x$, a student response $y$ produces a sequence of states
$s_t=(x,y_{<t})$, where $y_{<t}$ contains the response tokens preceding
position $t$. Different responses to the same query induce different
prefixes at which teacher supervision is applied. Current-policy rollouts
are generated by the student as it is updated; initial-policy rollouts are
generated by the student $\theta_0$ before OPD training. For a fixed query,
the policy that generates a rollout therefore determines the distribution
of prefixes at which supervision is applied.

\paragraph{Supervision.}
At each recorded state, the teacher's next-token probabilities supervise
the current student's predictions. Both models condition on the same query
and response prefix. Resampling provides new prefixes for training;
replaying stored responses returns to their existing prefixes while the
student continues to learn. Before each scoring batch, we recompute the
current student's and fixed teacher's token probabilities. This distinguishes
the source of training states from the current student's predictions at
those states. We use a reverse-KL-derived distillation objective, specified
in Appendix~\ref{app:protocol:objective}.

\paragraph{Training and evaluation.}
Our experiments vary the query set, rollout resampling schedule, and replay
schedule while controlling the trajectory and optimization-step budgets
within each comparison. The main analyses use Qwen3-0.6B as the student
and Qwen3-4B-Instruct-2507 as the teacher, with Qwen3-1.7B used to evaluate
transfer across student sizes \citep{yang2025qwen3}. We compute
response-level accuracy per question, average over questions within each
benchmark, and then average equally across benchmarks.
Appendices~\ref{app:protocol}, \ref{app:measurement}, and \ref{app:banks}
provide the full training configuration, evaluation procedures, and
query-set construction, respectively.

\section{Query Count and Rollout Sampling Interact}
\label{sec:grid}

OPD can obtain new training prefixes in two ways: by presenting the
student with different queries, or by resampling responses as the student
policy changes. A natural intuition is that more queries and more frequent
resampling should both improve the use of teacher supervision. We test
whether this intuition holds when the training budget is fixed.

\paragraph{Experimental design.}
We compare training with 8 queries, 48 queries, and the full
14,080-query training set, crossed with 1, 10, or 110 rollout collection
rounds. Every configuration consumes 14,080 trajectories over 110
optimization steps. With one collection round, all responses are generated
before training. With 10 rounds, a new block of responses is sampled every
11 optimization steps; with 110 rounds, responses are resampled before
every step. All configurations use the same distillation objective and a
16K training response limit. We evaluate on AIME24, AIME25, and AMC23 at
the same response limit, with $1\mathrm{K}=1{,}024$ tokens.

\begin{table}[!htbp]
\centering
\caption{\textbf{Query count and rollout sampling interact.}
Average 16K accuracy (\%) on AIME24, AIME25, and AMC23. All configurations
use 14,080 trajectories and 110 optimization steps. Per-benchmark results
and uncertainty estimates appear in Appendix~\ref{app:grid-details}.}
\label{tab:surface}
\begin{tabular}{lrrr}
\toprule
Training queries
& Initial rollouts only
& Resample every 11 steps
& Resample every step \\
\midrule
8                  & 21.16 & 24.09 & 23.61 \\
48                 & 20.18 & 24.40 & 24.22 \\
Full (14,080)      & 19.05 & 24.51 & 25.57 \\
\bottomrule
\end{tabular}
\end{table}

\paragraph{More queries and more frequent resampling are not monotonically better.}
The effect of adding queries depends on how responses are sampled.
With fixed initial-policy rollouts, expanding from 8 queries to the full
training set lowers average accuracy from $21.16\%$ to $19.05\%$.
With per-step resampling, the same change raises accuracy from $23.61\%$
to $25.57\%$ (Table~\ref{tab:surface}). The difference between these
effects is 4.07 percentage points ($95\%$ paired question-bootstrap
interval $[2.00,6.28]$).

More frequent resampling is likewise not uniformly beneficial. Increasing
the number of collection rounds from 10 to 110 lowers accuracy for the
8- and 48-query settings, but improves it for the full training set.
Moreover, most of the gain appears between one and ten collection rounds.
Query count and resampling frequency therefore interact under a fixed
trajectory budget.

\paragraph{Response limits affect the rollout comparison.}
The sampling schedules also produce markedly different response behavior.
With fixed initial-policy rollouts, only $21.2\%$--$24.9\%$ of responses
both contain an extractable answer and finish below the 16K limit,
compared with $92.6\%$--$96.8\%$ when rollouts are resampled. Roughly
half of the fixed-rollout responses reach the 16K cap without an
extractable answer.

The 16K comparison therefore mixes two effects: whether the model can
produce a correct answer, and whether that answer appears before the
response limit. A rollout source that appears worse at 16K may behave
differently when generation is allowed to continue. We therefore next
hold the query set fixed and ask whether the advantage of current-policy
rollouts persists as the response-length budget increases.

\section{The Value of Fresh Rollouts Depends on Response-Length Budget}
\label{sec:behavior}

Section~\ref{sec:grid} showed that rollout freshness is not uniformly
beneficial under a 16K response limit, and that different sampling schedules
produce markedly different termination behavior. We therefore ask whether
the advantage of current-policy rollouts persists when the model is allowed
to generate longer responses.

\paragraph{A matched comparison under two teachers.}
\label{sec:behavior:matched}
We train Qwen3-0.6B with either Qwen3-4B-Instruct-2507 or
JustRL-DeepSeek-1.5B (JustRL; \citealp{he2025justrl}). Within each teacher
comparison, the two conditions share the same initialization, 48 queries,
4,800 trajectories, 40 optimization steps, and a 16K training response
limit. One condition periodically resamples rollouts from the evolving
student; the other trains on responses generated before optimization begins.
Thus, the comparison changes the rollout source while holding the training
budget fixed.

On the eight-benchmark suite, current-policy sampling has higher average
accuracy at 16K, with larger gains in repeated-sampling coverage and answer
extraction
(Table~\ref{tab:rollout-budget-summary}).

\begin{table}[t]
\centering
\caption{\textbf{The relative value of rollout sources changes with response budget.}
Eight-benchmark average differences in percentage points: current-policy
minus initial-policy at 16K, and initial-policy minus current-policy
accuracy at 32K. Positive values favor the source named in the column
header. The 16K results re-score prefixes of the same 32K generations.
For pass@$n$, $n=32$ on the six competition benchmarks and $n=4$ on
MATH-500 and Olympiad-Bench. Extraction is the fraction of responses with
an extractable boxed answer.}
\label{tab:rollout-budget-summary}
\small
\setlength{\tabcolsep}{4pt}
\begin{tabular}{lrrrr}
\toprule
Teacher
& \multicolumn{3}{c}{\shortstack{Current-policy advantage\\at 16K}}
& \shortstack{Initial-policy advantage\\at 32K} \\
\cmidrule(lr){2-4}\cmidrule(l){5-5}
& Acc. & pass@$n$ & Extraction & Acc. \\
\midrule
Qwen3-4B
& +0.99 & +10.38 & +38.87 & \textbf{+4.39} \\
JustRL-1.5B
& +6.15 & +16.09 & +39.52 & \textbf{+2.07} \\
\bottomrule
\end{tabular}
\end{table}

\paragraph{The ranking reverses as the response budget increases.}
\label{sec:behavior:budget}
We evaluate each trained model on eight mathematics benchmarks with a 32K
generation limit. We re-tokenize the saved responses with the student
tokenizer and re-score their first 4K, 8K, 16K, or 24K tokens.
This measures when correct answers become available within the same sampled
continuations.

Current-policy training retains higher average accuracy through 16K under both
teachers. The ordering then reverses: between 16K and 24K with Qwen3, and
between 24K and 32K with JustRL
(Figure~\ref{fig:budgetcrossover}a). At 32K, initial-policy training leads
by 4.39 percentage points with Qwen3 ($95\%$ interval $[3.07,5.75]$) and
2.07 points with JustRL ($[0.49,3.67]$). The advantage appears on all eight
benchmarks with Qwen3 and on six of eight with JustRL, with one tie.

The reversal is accompanied by longer responses under initial-policy
training: mean response length at 32K is 21,736 versus 12,011 tokens with
Qwen3, and 24,359 versus 14,152 with JustRL.

\section{What Remains as Supervision Moves On?}
\label{sec:initial-states}
\label{sec:replay:states}

The longer-budget advantage of initial-policy rollouts raises a further
question: do their prefixes still offer useful teacher supervision after
current-policy training?

\paragraph{Fixing prefixes separates state visitation from prediction.}
For each of the same 48 queries, the initial student generates eight
diagnostic responses. We sample 32 positions per response and evaluate the
same prefixes at initialization and after 20 and 40 optimization steps on
current-policy rollouts, denoting the latter checkpoints by $R_{20}$ and
$R_{40}$.

For a recorded prefix $y_{<t}$, $\pi_\theta(y_{<t}\mid x)$ measures how
likely the student is to generate that exact prefix. At the fixed state
$(x,y_{<t})$, reverse KL between the student's and teacher's full-vocabulary
next-token distributions measures their conditional disagreement.

Evaluating the same prefixes across checkpoints separates changes in state
visitation from changes in the student's conditional predictions.

\begin{figure}[!ht]
\centering
\includegraphics[width=\textwidth]{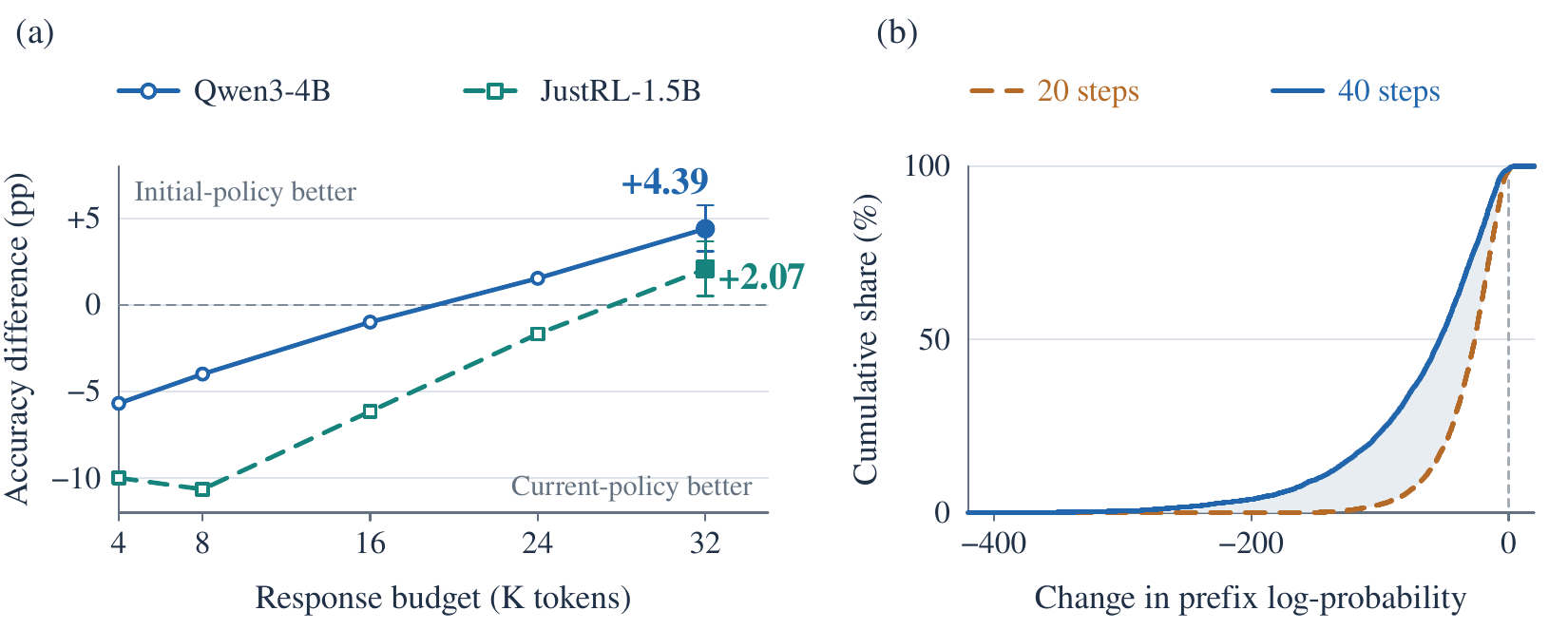}
\caption{\textbf{Rollout-source advantages depend on response budget,
while earlier prefixes become less likely.}
(a) Initial-policy minus current-policy accuracy after 40 optimization
steps, averaged equally over eight benchmarks under two teachers.
Budgets re-score saved 32K responses; error bars show 95\% paired-bootstrap
intervals at 32K.
(b) Response-weighted cumulative distributions of the change in initial-policy
prefix log-probability from initialization to 20 steps (orange dashed) or
40 steps (blue solid). Negative changes mean the prefixes become less likely;
the light fill marks the gap between the curves. Each response has equal
weight; see Appendix~\ref{app:replay:cross-scoring} for sampling details.}
\label{fig:budgetcrossover}
\label{fig:prefix-diagnostics}
\end{figure}

\paragraph{Earlier prefixes become less likely even as predictions improve.}
After 40 optimization steps, 99.15\% of sampled nonempty initial-policy
prefixes are less likely than at initialization, with responses weighted
equally (Figure~\ref{fig:prefix-diagnostics}b). Mean teacher--student KL
on the initial-policy prefixes decreases from 0.774 at initialization to
0.548 after 20 steps and 0.448 after 40
(Table~\ref{tab:rf-cross-kl}).

The student thus becomes less likely to revisit these exact prefixes even
though its predictions there remain different from the teacher's. We next
test whether replaying these states improves accuracy after current-policy
training.

\section{Revisiting Earlier Student States}
\label{sec:replay}

Section~\ref{sec:initial-states} found residual teacher--student disagreement
on earlier prefixes after on-policy training. We now test whether replaying
these states improves accuracy and whether the training order matters.
All comparisons use the same 48 queries, Qwen3-0.6B student, and
Qwen3-4B-Instruct-2507 teacher.

\subsection{Which trajectories should be replayed?}
\label{sec:replay:switch}
\label{sec:replay:source}

From the same $R_{40}$ model and optimizer checkpoint, we add 20 optimization
steps using one of three sources: fresh current-policy rollouts, a fixed
initial-policy rollout bank, or a fixed bank newly generated from $R_{40}$.
All endpoints consume 7,200 trajectories over 60 optimization steps.
The fixed recent bank tests whether holding response text fixed can
reproduce the benefit of returning to initial-policy prefixes.

\begin{table}[!htbp]
\centering
\caption{\textbf{The source and order of replay affect accuracy.}
Eight-benchmark mean accuracy (\%) for Qwen3-0.6B. Source compares 20-step
continuations after 40 current-policy steps; the fixed recent bank comes
from the 40-step student. Order compares two 20-step stages after 20
current-policy steps. All endpoints use 7,200 trajectories and 60 updates.
16K re-scores saved 32K responses; Tokens is mean 32K output length.
Bold marks the best accuracy within each comparison.}
\label{tab:state-source-core}
\label{tab:state-order-core}
\small
\setlength{\tabcolsep}{5pt}
\begin{tabular}{@{}llrrr@{}}
\toprule
Comparison & Schedule & 16K & 32K & Tokens \\
\midrule
\multirow{3}{*}{Source} & Current-policy sampling & 26.41 & 26.52 & 8,254 \\
 & Fixed recent bank & 26.12 & 26.21 & 7,997 \\
 & Initial-policy replay & \textbf{27.13} & \textbf{29.37} & 12,116 \\
\midrule
\multirow{2}{*}{Order} & Current then initial & \textbf{27.77} & \textbf{30.55} & 13,015 \\
 & Initial then current & 26.16 & 27.24 & 11,369 \\
\bottomrule
\end{tabular}
\end{table}

Initial-policy replay improves average accuracy over continued current-policy
sampling at both 16K and 32K, with a larger gain at 32K
(Table~\ref{tab:state-source-core}). Fixed recent rollouts remain close to
continued sampling, indicating that the gain depends on the trajectory
source rather than freezing responses alone. Both banks contain 2,400
responses from the same 48 queries: initial-policy responses average
758.85 tokens, versus 8,171.47 for the $R_{40}$ bank. The shorter initial
responses yield higher accuracy, so longer replayed answers alone cannot
explain this advantage. These are training-bank lengths; the table's Tokens
column reports final-model evaluation outputs, which are longer after replay.

Initial-policy replay reduces KL more on initial-policy prefixes, whereas
continued current-policy sampling reduces it more on recent $R_{40}$
prefixes (Table~\ref{tab:replay-kl-core}). Thus, the rollout source
changes which prefixes receive the larger improvement in teacher alignment.

\begin{table}[!htbp]
  \centering
  \caption{\textbf{Rollout source determines where KL decreases.}
  Full-vocabulary reverse KL (nats; lower is better) on fixed diagnostic
  prefixes generated by the initial and 40-step students. All continuations
  start from the same $R_{40}$ checkpoint and add 20 optimization steps.
  Responses are weighted equally; the full matrix is in
  Table~\ref{tab:rf-cross-kl}.}
  \label{tab:replay-kl-core}
  \small
  \begin{tabular}{lrr}
    \toprule
    Student checkpoint / continuation & Initial-policy prefixes & $R_{40}$ prefixes \\
    \midrule
    Common starting checkpoint ($R_{40}$) & 0.448 & 0.427 \\
    \midrule
    Continue current-policy sampling & 0.404 & 0.370 \\
    Replay fixed recent rollouts & 0.404 & 0.370 \\
    Replay initial-policy rollouts & 0.370 & 0.400 \\
    \bottomrule
  \end{tabular}
\end{table}

\subsection{When should these trajectories be replayed?}
\label{sec:replay:order}

From the same $R_{20}$ model and optimizer checkpoint, we apply 20
current-policy steps followed by 20 initial-policy replay steps, or reverse
these two stages. Both orders use the same initial-policy bank and finish
with 40 current-policy and 20 replay steps in total. Current-policy rollouts
are generated at their assigned stage.

Current-policy training followed by replay outperforms the reverse order
by 1.61 percentage points at 16K and 3.31 at 32K, with gains on all eight
benchmarks (Table~\ref{tab:state-order-core}). The reverse order remains
more accurate at 4K and 8K (Appendix~\ref{app:replay:controls}), so the
preferred order depends on the response budget.

\section{\ropd{}: Revisiting Initial-Policy Rollouts}
\label{sec:annealing}

Section~\ref{sec:replay} shows that current-policy learning before replay
outperforms the reverse order at 16K and 32K. This suggests a two-stage
curriculum: first train on the student's evolving states, then revisit
initial states that still carry teacher supervision. The key decision is
when to start replay. Rather than prescribe the same switch iteration for
every run, \ropd{} detects when current-policy gradients stabilize relative
to their variation within an iteration.

\subsection{Adaptive replay curriculum}
Before training, we save 2,400 responses sampled from the initial student
on the same 48 queries. Each iteration uses 480 responses from one source
for all four optimizer minibatches. Initially, the student generates fresh
current-policy rollouts. If the trigger fires at iteration $\tau$, that
iteration finishes on current-policy rollouts; from iteration $\tau+1$
onward, training uses only the saved initial-policy responses. Replay
reuses their text while recomputing student and teacher token probabilities
for the distillation loss.
Let $\alpha_k\in\{0,1\}$ denote this source choice, with $0$ for current-policy
sampling and $1$ for initial-policy replay.

\subsection{Gradient-triggered transition}
\label{sec:annealing:trigger}
Before replay, let $g_{k,1},\ldots,g_{k,m}$ be the pre-clipping,
pre-AdamW gradients of iteration $k$ with $m=4$ minibatches. Define
\begin{equation}
\bar g_k=\frac{1}{m}\sum_{i=1}^{m}g_{k,i},
\qquad
v_k=\frac{1}{m(m-1)}
\sum_{i=1}^{m}\|g_{k,i}-\bar g_k\|_2^2.
\label{eq:gradient-variation}
\end{equation}
Let $p(k)$ be the preceding current-policy iteration. We define
\begin{equation}
D_k=\|\bar g_k-\bar g_{p(k)}\|_2^2,
\qquad
V_k=v_k+v_{p(k)}.
\label{eq:gradient-drift}
\end{equation}
$D_k$ measures how much the mean gradient changes between current-policy
iterations; $V_k$ measures its variation across minibatches in those two
iterations. Because parameters change between minibatches, $V_k$ serves as
an empirical comparison scale. When $D_k\leq V_k$, the change across
iterations is no larger than this within-iteration scale. We use two
consecutive qualifying comparisons to mark the end of the current-policy
stage:
\begin{equation}
D_k \leq V_k.
\label{eq:anneal-trigger}
\end{equation}
If the trigger occurs at iteration $\tau$, the source switches at the
next iteration and remains on initial-policy replay:
\begin{equation}
\alpha_k=
\begin{cases}
0, & k\leq\tau,\\[2pt]
1, & \tau<k\leq K,
\end{cases}
\label{eq:anneal-schedule}
\end{equation}
where $K$ is the total number of training iterations. If no trigger occurs,
$\alpha_k$ remains zero.

\begin{figure}[!ht]
\centering
\includegraphics[width=\textwidth]{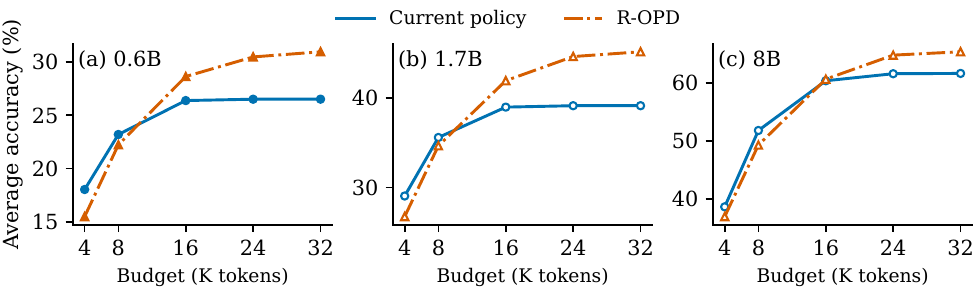}
\caption{\textbf{\ropd{} across response budgets.}
Eight-benchmark accuracy after 60 updates: current-policy sampling versus
\ropd{}. Curves average three runs for 0.6B and show one run for 1.7B/8B;
Table~\ref{tab:annealing-main}a summarizes a separate three-data-order set
at every size.
Shorter budgets re-score saved 32K responses. Teachers:
Qwen3-4B-Instruct-2507 in (a,b), Qwen3-30B-A3B-Instruct-2507 in (c).}
\label{fig:annealing-schedule}
\label{fig:curriculum-profile}
\par\vspace{6pt}
\begingroup
\makeatletter\def\@captype{table}\makeatother
\centering
\caption{\textbf{Fixed and gradient-triggered replay.}
Eight-benchmark accuracy (\%) across three orders of 48 queries.
(a) Mean $\pm$ sample SD ($n=3$); Tokens are rounded mean 32K output lengths
from the same runs. Fixed $R_{40}F_{20}$ uses 40 current-policy updates,
then 20 initial-policy replay updates. (b) 0.6B fixed switch iteration
$\tau$ versus \ropd{} triggers $(5,5,7)$; $\tau=0$ uses initial-policy
replay throughout and $\tau=15$ uses current-policy sampling throughout.
Each run has 15 iterations. Teachers: Qwen3-4B-Instruct-2507 for 0.6B/1.7B,
Qwen3-30B-A3B-Instruct-2507 for 8B.}
\label{tab:annealing-main}
\label{tab:annealing-transfer}
\label{tab:annealing-fixed-onset}
\small
\makebox[\linewidth][l]{\textit{(a) Results across student sizes}}\par\vspace{2pt}
\begin{tabular}{llrrr}
\toprule
Student & Method & 16K accuracy (\%) & 32K accuracy (\%) & Tokens at 32K \\
\midrule
0.6B & Current policy & $26.37\,\pm\,0.41$ & $26.50\,\pm\,0.42$ & 8,328 \\
 & Fixed R40F20 & $27.38\,\pm\,0.22$ & $29.56\,\pm\,0.16$ & 12,036 \\
 & \ropd{} & $28.62\,\pm\,0.30$ & $30.94\,\pm\,0.53$ & 12,617 \\
\midrule
1.7B & Current policy & $39.07\,\pm\,0.40$ & $39.18\,\pm\,0.35$ & 6,713 \\
 & Fixed R40F20 & $41.19\,\pm\,0.09$ & $43.67\,\pm\,0.06$ & 10,933 \\
 & \ropd{} & $41.88\,\pm\,0.21$ & $45.33\,\pm\,0.16$ & 11,589 \\
\midrule
8B & Current policy & $60.32\,\pm\,0.17$ & $61.57\,\pm\,0.08$ & 7,866 \\
 & Fixed R40F20 & $60.23\,\pm\,0.31$ & $63.77\,\pm\,0.25$ & 10,419 \\
 & \ropd{} & $61.10\,\pm\,0.40$ & $65.62\,\pm\,0.22$ & 11,121 \\
\bottomrule
\end{tabular}

\par\vspace{7pt}
\makebox[\linewidth][l]{\textit{(b) Replay-onset scan for Qwen3-0.6B}}\par\vspace{2pt}
\begingroup
\setlength{\tabcolsep}{3pt}
\begin{tabular*}{\linewidth}{@{}l@{\extracolsep{\fill}}rrrrrrrrrr@{}}
\toprule
$\tau$ & 0 & 4 & 5 & 6 & 7 & 8 & 10 & 12 & 15 & \ropd{} \\
\midrule
16K & 25.26 & 27.75 & 28.46 & 28.45 & 28.36 & 27.99 & 27.38 & 26.84 & 26.37 & 28.62 \\
32K & 30.31 & 30.68 & 30.79 & 30.74 & 30.64 & 30.42 & 29.56 & 28.17 & 26.50 & 30.94 \\
\bottomrule
\end{tabular*}
\endgroup

\endgroup
\end{figure}

\subsection{Results across student sizes and response budgets}
\label{sec:annealing:results}
\label{sec:annealing:transfer}
Table~\ref{tab:annealing-main} compares \ropd{}, current-policy sampling,
and fixed $R_{40}F_{20}$ over three data orders of 48 queries, with a 16K
training response limit and 15 iterations (60 updates; 7,200 trajectories).
The 0.6B/1.7B students share the 4B teacher and trigger parameters;
each uses its own initial-policy bank and measured trigger.

For Qwen3-0.6B, \ropd{} improves mean accuracy over continued
current-policy sampling by 2.25 points at 16K and 4.44 points at 32K.
For Qwen3-1.7B, the corresponding mean gains are 2.81 and 6.15 points
(Table~\ref{tab:annealing-main}).
Against fixed $R_{40}F_{20}$, \ropd{} gains 1.24 points at 16K and
1.39 at 32K for 0.6B. For 1.7B, \ropd{} gains 0.68 and
1.66 points, respectively.

With Qwen3-8B and a Qwen3-30B-A3B-Instruct-2507 teacher, three paired
runs yield mean 32K gains of 2.20/4.05 points for fixed replay/\ropd{}
over current-policy sampling (Table~\ref{tab:annealing-main}). At 16K,
\ropd{} gains 0.78 points over current-policy sampling and exceeds fixed
replay by 0.87 points; its 32K advantage over fixed replay is 1.85 points.

Across three training-data shuffle seeds per size, the trigger iteration
is $5.67\pm1.15$, $6.67\pm0.58$, and $7.67\pm0.58$ for 0.6B, 1.7B,
and 8B, respectively (mean $\pm$ sample SD;
Appendix~\ref{app:annealing:repeats}). The trigger thus varies with
student size and data order.

For 0.6B, a three-shuffle fixed-onset scan finds that $\tau=5$ reaches 28.46/30.79\%
(16K/32K), close to \ropd{}'s 28.62/30.94\% with triggers $(5,5,7)$,
whereas $\tau=10$ yields 27.38/29.56\%
(Table~\ref{tab:annealing-main}b). Earlier replay explains part of
the original baseline gap. \ropd{} selects onsets in the high-performing
region of the scan and attains accuracy close to the best fixed setting
without prescribing a switch iteration for each run.

Continued current-policy sampling remains stronger at 4K and 8K, whereas
\ropd{} is more accurate at 16K and above
(Figure~\ref{fig:curriculum-profile}a,b). At the same evaluation prompts,
decoding settings, and 32K cap, \ropd{} produces higher accuracy and
52\%/73\%/41\% longer mean responses for 0.6B/1.7B/8B. The student
chooses these longer trajectories itself; the generation budget is
unchanged. Together with the 16K gains, this pattern suggests that
revisiting initial states at the selected time helps the student express
more of its reasoning capacity.

\section{Related Work}
\label{sec:related}

OPD trains on student states \citep{gu2024minillm,agarwal2024gkd}.
Analyses study token corrections and alignment
\citep{armandpour2026unmasking,zhu2026manyfaces,li2026rethinkingopd};
OPD variants alter probability updates or prefixes
\citep{xia2026routeopd,zhang2026prefixopd}, while RL work selects
informative samples and tokens \citep{wang2026d3s}.
We hold the per-state objective and candidate selection fixed while varying
rollout origin, response budget, and training order.

Concurrent work explains few-query OPD through broad state coverage and
slow alignment \citep{fu2026onetrainingexample}. We study how the value of
student trajectories depends on their policy of origin and when to revisit
earlier states. Trajectory reuse and rollout freshness are also studied in
\citet{arnal2026replay,chen2026fopd}.
Our initial-policy bank stores response text; student and teacher
probabilities are recomputed for each scoring batch. Rollout age therefore
determines the state source without making the update probabilities stale.
Appendix~\ref{app:related} covers broader connections to reasoning
post-training, supervision design, and compact training sets.

\section{Conclusion}
\label{sec:conclusion}

OPD students determine which prefixes receive teacher supervision.
More queries or fresher rollouts do not uniformly help; earlier states
can retain useful supervision after becoming less likely.
Across eight mathematics benchmarks and three data orders per student size,
\ropd{} selects when to revisit those states and exceeds current-policy
training and fixed $R_{40}F_{20}$ replay at 16K/32K for three Qwen3 student
sizes. Its higher accuracy and longer self-generated responses at 32K
suggest that timely replay helps bring out more of the student's reasoning
capacity.

\clearpage
\section*{AI Use Statement}

Generative AI tools assisted with manuscript polishing, related-work retrieval,
and debugging portions of the experimental code. The authors reviewed these
contributions and take responsibility for the paper and its artifacts.

\bibliography{iclr2027_conference}
\bibliographystyle{iclr2027_conference}

\appendix

\section{Training Protocol}
\label{app:protocol}

\subsection{Rollouts, training units, and experiment budgets}
\label{app:protocol:units}

We use \emph{rollout} for a sampled trajectory and \emph{response} for its
generated text. Current-policy rollouts are sampled from the evolving student;
the initial-policy bank is generated by $\theta_0$ before OPD training. Both
include responses regardless of correctness. Replay fixes the stored response
prefixes while supervising the current student.

An \emph{optimization step} is one AdamW update. A \emph{training iteration}
scores a rollout batch and applies one or more such updates; a \emph{rollout
collection round} samples trajectories from one student checkpoint. We use
$N$, $M$, $C$, and $U$ for the numbers of distinct queries, collection rounds,
consumed trajectories, and optimization steps, respectively. Replayed
trajectories count toward $C$. Comparisons match $C$ and $U$; token consumption
is reported separately because it also depends on response length.

Table~\ref{tab:experiment-map} maps the budgets to the query grid in
Section~\ref{sec:grid}, the source comparison in Section~\ref{sec:behavior},
the replay controls in Section~\ref{sec:replay}, and the adaptive and fixed
curricula in Section~\ref{sec:annealing}. Appendix~\ref{app:banks}
specifies query selection and trajectories per query.

\begin{table}[ht]
  \centering
  \small
  \setlength{\tabcolsep}{5pt}
  \caption{\textbf{Experiment budgets and evaluation protocols.}
  Each comparison matches consumed trajectories $C$ and optimization steps
  $U$. Three-benchmark evaluation uses direct 16K generations; eight-benchmark
  evaluation uses saved 32K responses and their shorter prefixes. The matched
  rollout comparison reports both protocols.}
  \label{tab:experiment-map}
  \begin{tabular}{lrrrl}
    \toprule
    Experiment & Queries & $C$ & $U$ & Evaluation \\
    \midrule
    Query--sampling grid & 8 / 48 / Full & 14,080 & 110 & Three benchmarks \\
    Matched rollout sources & 48 & 4,800 & 40 & Three and eight benchmarks \\
    Source and order & 48 & 7,200 & 60 & Eight benchmarks \\
    \ropd{} (0.6B, 1.7B, 8B) & 48 & 7,200 & 60 & Eight benchmarks \\
    Fixed-onset scan (0.6B) & 48 & 7,200 & 60 & Eight benchmarks \\
    \bottomrule
  \end{tabular}
\end{table}

\subsection{Distillation objective and shared configuration}
\label{app:protocol:objective}

Let \(\pi_\theta\) be the student, \(\pi_T\) the fixed teacher, and
\(s_t=(x,y_{<t})\) a state consisting of a query and a student response
prefix. For each state, the implementation stores the set \(\mathcal C_t\) of
the student's 16 highest-probability tokens. Define full-vocabulary log probabilities
\[
  \ell^S_{tj}=\log\pi_{\mathrm{old}}(j\mid s_t),\qquad
  \ell^T_{tj}=\log\pi_T(j\mid s_t),\qquad j\in\mathcal C_t,
\]
and the student mass renormalized only for weighting,
\[
  \bar p^S_{tj} =
  \frac{\exp(\ell^S_{tj})}
       {\sum_{k\in\mathcal C_t}\exp(\ell^S_{tk})}.
\]
The coefficient for candidate token \(j\) at position \(t\) is defined with
gradients stopped:
\[
  A_{tj}=\operatorname{sg}\!\left[
    \bar p^S_{tj}\bigl(\ell^T_{tj}-\ell^S_{tj}\bigr)
  \right].
\]
For \(r_{tj}(\theta)=\pi_\theta(j\mid s_t)/
\pi_{\mathrm{old}}(j\mid s_t)\), the implemented dual-clipped PPO term is
\[
\phi(A,r)=
\begin{cases}
\max\{-Ar,-A\,\mathrm{clip}(r,0.8,1.2)\}, & A\geq 0,\\
\min\{-3A,\max[-Ar,-A\,\mathrm{clip}(r,0.8,1.2)]\}, & A<0.
\end{cases}
\]
For a microbatch \(\mathcal B\), the actor loss is
\begin{equation}
  \mathcal L_{\mathrm{OPD}}(\theta;\mathcal B)=
  \frac{1}{\sum_{t\in\mathcal B} m_t}
  \sum_{t\in\mathcal B} m_t\sum_{j\in\mathcal C_t}
  \phi\!\left(A_{tj},r_{tj}(\theta)\right),
  \label{eq:implemented-opd}
\end{equation}
where \(m_t\) masks padding. Microbatch losses are weighted by sequence count
before gradient accumulation. Before each training iteration, both models
score the recorded prefixes by teacher forcing. The current student's
candidate sets, coefficients, and denominator policy \(\pi_{\mathrm{old}}\) are
then fixed for that iteration, including for replayed responses. Thus, the
loss is a reverse-KL-derived surrogate over the student's top-16 tokens:
log probabilities use the full vocabulary, while student weights are
normalized on \(\mathcal C_t\).

All comparisons share this objective and candidate rule. The implementation
uses \texttt{top\_k=16}, \texttt{top\_k\_strategy=only\_stu}, and
\texttt{reward\_weight\_mode=student\_p}. Task rewards, verifier scores,
selective-token masks, entropy bonuses, and auxiliary KL losses are disabled.
Dynamic microbatches accommodate long contexts across eight FSDP ranks.

\begin{table}[ht]
  \centering
  \small
  \caption{\textbf{Shared training configuration.}
  Training queries come from the filtered English subset of DAPO-Math-17K
  \citep{yu2025dapo}; Appendix~\ref{app:banks} gives the construction.}
  \label{tab:setup-config}
  \begin{tabular}{ll}
    \toprule
    Main student & Qwen3-0.6B, non-thinking \\
    Curriculum-transfer students & Qwen3-1.7B and Qwen3-8B, non-thinking \\
    8B curriculum teacher & Qwen3-30B-A3B-Instruct-2507 \\
    Main teacher & Qwen3-4B-Instruct-2507 (shared tokenizer) \\
    Secondary teacher & JustRL-DeepSeek-1.5B (cross-tokenizer) \\
    Actor/ref parameters & FP32 \\
    Teacher scoring & FP32 parameters, BF16 autocast \\
    Rollout engine & vLLM, BF16 \\
    Optimizer & AdamW, $\beta=(0.9,0.999)$, weight decay $0.01$ \\
    Learning-rate schedule & constant $10^{-6}$, no warmup \\
    Gradient clipping & global norm $1.0$ \\
    PPO configuration & one epoch, minibatch 128, clip $0.2$, dual clip $3.0$ \\
    Loss aggregation & sum over candidates, mean over valid response tokens \\
    Max training response / prompt length & 16,384 / 1,024 tokens \\
    Rollout sampling & temperature $1.0$, top-$p$ $1.0$, one response per query \\
    Parallelism & 8 FSDP ranks \\
    \bottomrule
  \end{tabular}
\end{table}

Data order stays fixed within a run: the trainer does not reshuffle it.
For the three data-order repetitions in Section~\ref{sec:annealing},
the same 48-query set is permuted before each run and then held fixed.
Evaluation uses the checkpoint at the specified final optimization step.

\subsection{Rollout sampling schedules}
\label{app:protocol:sampling}

For the central grid, every configuration uses $C=14{,}080$ and $U=110$:
\begin{itemize}
  \item $M=1$: all 14,080 responses are generated from $\theta_0$ before
    training and partitioned into 110 consecutive optimizer minibatches
    of 128 trajectories. Each trajectory is consumed once.
  \item $M=10$: each of ten rollout collection rounds generates 1,408 responses from the current
    student model, then partitions them into 11 optimizer minibatches of 128.
  \item $M=110$: before every optimization step, 128 responses are generated from the current
    student. This is the per-step rollout sampling condition.
\end{itemize}
For the matched $C=4{,}800$, $U=40$ comparison in
Section~\ref{sec:behavior:matched}, periodic sampling uses ten iterations of
480 responses, each consumed in steps of sizes $128/128/128/96$. The
initial-policy condition generates all 4,800 responses at $\theta_0$ and uses
the same 40-step partition. The paired conditions otherwise share queries,
initialization, objective, and update schedule.

For the cross-tokenizer teacher, student candidate strings are mapped to teacher
token IDs before teacher scoring. Candidates without a single-token mapping are
masked from the support. Eight Qwen3 control tokens are unmappable under this
rule; the semantic EOS mapping is retained. This comparison changes the
teacher, its training history, and tokenizer, and therefore serves only as a
replication of the rollout-source effect.

\section{Measurement Details}
\label{app:measurement}

\subsection{Evaluation suites and response budgets}
\label{app:measurement:suites}

The three-benchmark suite contains AIME24, AIME25, and AMC23 (143 questions;
4,576 responses per checkpoint). The eight-benchmark suite contains every row
of Table~\ref{tab:evaluation-suites} (1,411 questions; 12,252 responses).
Metrics are computed per benchmark and then averaged equally within the suite.

\begin{table}[ht]
  \centering
  \small
  \caption{\textbf{Evaluation sets and sampling counts per checkpoint.}}
  \label{tab:evaluation-suites}
  \begin{tabular}{lrrr}
    \toprule
    Benchmark & Questions & Responses per question & Total responses \\
    \midrule
    AIME24 & 30 & 32 & 960 \\
    AIME25 & 30 & 32 & 960 \\
    AIME26 & 30 & 32 & 960 \\
    AMC23 & 83 & 32 & 2,656 \\
    HMMT25-Nov & 30 & 32 & 960 \\
    HMMT26-Feb & 33 & 32 & 1,056 \\
    MATH-500 & 500 & 4 & 2,000 \\
    OlympiadBench & 675 & 4 & 2,700 \\
    \bottomrule
  \end{tabular}
\end{table}

Evaluation uses temperature $1.0$, top-$p$ $1.0$, and non-thinking mode. The
query--sampling grid and three-benchmark matched-source results use direct 16K
generation. Eight-benchmark evaluation generates to 32K and re-scores the same
responses at 4K, 8K, 16K, 24K, and 32K under the student tokenizer, where
$1\mathrm{K}=1{,}024$ tokens. Direct 16K generations and 16K prefixes of 32K
generations are separate Monte Carlo samples.

\subsection{Metrics and response categories}
\label{app:measurement:metrics}

\paragraph{Accuracy, pass, and majority vote.}
For question $i$ with $n_i$ responses and $c_i$ correct responses,
response-level accuracy is $c_i/n_i$, averaged over questions for benchmark
accuracy. Pass@$n$ is the fraction of questions with at least one correct
response; majority-vote accuracy uses the most frequent extracted answer.
Here $n=32$ for the six competition benchmarks and $n=4$ for MATH-500 and
OlympiadBench.

\paragraph{Answer extraction, length, and truncation.}
The evaluator records correctness and whether a boxed answer can be extracted;
the latter defines \emph{Extraction} in tables. All conditions use the same
version of \texttt{ttrl\_math.compute\_score}. A response can contain an
extractable or correct answer before the cutoff even if generation reaches it.

\begin{figure}[H]
  \centering
  \includegraphics[width=\textwidth]{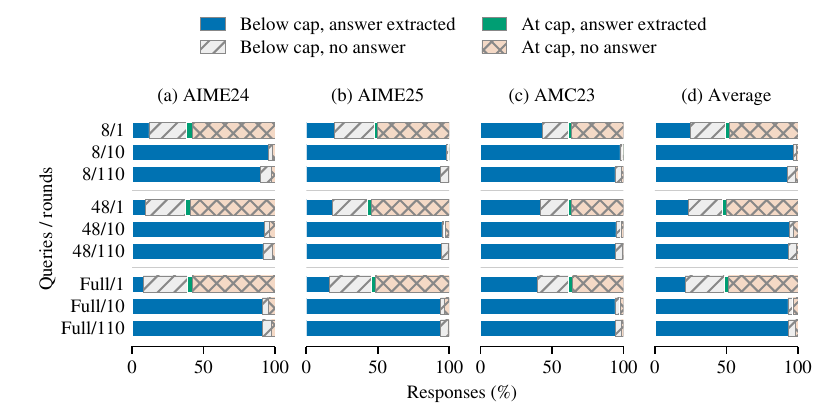}
  \caption{\textbf{Response categories by benchmark and sampling configuration.}
  Labels give the query count and collection rounds. Bars show the four mutually
  exclusive combinations of answer extraction and reaching the 16K limit;
  extraction counts answers regardless of correctness. The average weights
  the three benchmarks equally.}
  \label{fig:validity}
\end{figure}

Mean response length is the retained token count averaged under the student
tokenizer. In Figure~\ref{fig:validity}, ``at cap'' means that retokenized
length reaches the nominal limit; responses are partitioned by this indicator
and whether an answer is extracted. Category accuracy and overall accuracy are
computed separately.

\subsection{Uncertainty}
\label{app:measurement:uncertainty}

Intervals in Table~\ref{tab:refresh} use 10,000 draws that resample matched
questions within each benchmark and average the three benchmark differences.
The interaction in Equation~\ref{eq:interaction} uses 20,000 paired draws
across its four cells (seed 20260906). The 32K differences in
Section~\ref{sec:behavior} use 20,000 analogous draws across eight benchmarks
(seed 20260907).

These percentile 95\% intervals quantify question-level uncertainty for fixed
models and stored samples. Budget-curve intervals are pointwise at 32K; earlier
prefixes have point estimates. The curriculum comparison in Section~\ref{sec:annealing} reports
sample standard deviations over three data-order runs per method and
student size on the same fixed 48-query set.
The continuation results in Section~\ref{sec:replay} are descriptive. Gradient
uncertainties are standard errors over 16 diagnostic minibatches unless stated
otherwise.

\raggedbottom
\makeatletter
\setlength{\@fptop}{0pt}
\setlength{\@fpsep}{16pt}
\setlength{\@fpbot}{0pt plus 1fil}
\makeatother
\section{Training Query Sets}
\label{app:banks}

\paragraph{Source corpus.}
All queries come from the English split of DAPO-Math-17K, containing
$14{,}116$ raw rows. Filtering with the Qwen3-0.6B chat template and a
$1{,}024$-token prompt limit leaves $14{,}107$ usable rows. The full
training set contains the first $14{,}080$ of these rows, matching the
trajectory budget $\Ctraj$ of the query--sampling grid.

\paragraph{Query selection.}
The random 48-query set uses seed \texttt{20260829}, without consulting
rewards, difficulty labels, or model outputs. The eight-query set is a
strict subset of this set. The difficulty-selected 48-query set
in Table~\ref{tab:anchors} combines two selection rounds of 25 and 34
queries with 11 in common. It is the only query set selected using model
behavior.

\paragraph{Trajectories per query.}
At $C=14{,}080$, the eight-query, 48-query, and full-set conditions use
1,760, approximately 293, and one trajectory per query, respectively.
At $C=4{,}800$ with 48 queries, each query has 100 trajectories.

\paragraph{Effect of query selection (Section~\ref{sec:grid}).}
Table~\ref{tab:anchors} compares random and difficulty-selected query sets
under the same training budget. Random selection gives similar accuracy
and observed pass@32, supporting the use of the random set in the main
experiments.
\begin{table}[!htbp]
\centering\small
\caption{\textbf{Query selection at a matched training budget.}
Qwen3-0.6B with the 4B teacher; all conditions use 14,080 trajectories,
110 updates, and per-step resampling. Accuracy and observed pass@32 (\%)
are equal-weight averages over AIME24, AIME25, and AMC23 at a direct 16K limit.}
\label{tab:anchors}
\begin{tabular}{lrrr}
\toprule
Query set & Queries & Accuracy & pass@32 \\
\midrule
Full & 14,080 & 25.57 & 58.11 \\
Difficulty-selected & 48 & 23.98 & 58.51 \\
Random & 48 & 24.22 & 59.22 \\
\bottomrule
\end{tabular}
\end{table}
\FloatBarrier

\section{Query-Count and Rollout-Source Results}
\label{app:additional}

This appendix expands the query-count interaction in Section~\ref{sec:grid}
and the rollout-source reversal in Section~\ref{sec:behavior}.

\subsection{Query count and rollout sampling}
\label{app:grid-details}

All nine configurations use 14,080 trajectories, 110 optimization steps, and
a 16K training response limit (Appendix~\ref{app:protocol:sampling}). Table~\ref{tab:grid-perdataset}
reports accuracy, observed pass@32, and extraction from separate 16K
evaluations on three benchmarks; each configuration is one training run.

\begin{table}[H]
\centering
\caption{\textbf{Per-benchmark results for the query-count and sampling grid} (\%). Avg. weights AIME24, AIME25, and AMC23 equally; $N$ and $M$ denote query and collection-round counts.}
\label{tab:grid-perdataset}
\label{tab:surface-full}
\small
\setlength{\tabcolsep}{5pt}
\begin{tabular}{@{}llrrrrr@{}}
\toprule
Metric & $N$ & $M$ & AIME24 & AIME25 & AMC23 & Avg. \\
\midrule
\multirow{9}{*}{Accuracy} & 8 & 1 & 8.85 & 14.90 & 39.72 & 21.16 \\
 & 8 & 10 & 11.46 & 15.42 & 45.41 & 24.09 \\
 & 8 & 110 & 10.83 & 15.42 & 44.58 & 23.61 \\
 & 48 & 1 & 7.29 & 14.90 & 38.37 & 20.18 \\
 & 48 & 10 & 11.98 & 15.31 & 45.90 & 24.40 \\
 & 48 & 110 & 12.08 & 15.10 & 45.48 & 24.22 \\
 & Full & 1 & 6.35 & 13.65 & 37.16 & 19.05 \\
 & Full & 10 & 12.40 & 15.00 & 46.12 & 24.51 \\
 & Full & 110 & 13.12 & 16.04 & 47.55 & 25.57 \\
\midrule
\multirow{9}{*}{Observed pass@32} & 8 & 1 & 40.00 & 36.67 & 62.65 & 46.44 \\
 & 8 & 10 & 43.33 & 33.33 & 81.93 & 52.87 \\
 & 8 & 110 & 53.33 & 40.00 & 81.93 & 58.42 \\
 & 48 & 1 & 36.67 & 26.67 & 61.45 & 41.59 \\
 & 48 & 10 & 50.00 & 50.00 & 84.34 & 61.45 \\
 & 48 & 110 & 53.33 & 40.00 & 84.34 & 59.22 \\
 & Full & 1 & 30.00 & 30.00 & 59.04 & 39.68 \\
 & Full & 10 & 50.00 & 40.00 & 83.13 & 57.71 \\
 & Full & 110 & 46.67 & 43.33 & 84.34 & 58.11 \\
\midrule
\multirow{9}{*}{Extraction} & 8 & 1 & 16.15 & 21.77 & 45.90 & 27.94 \\
 & 8 & 10 & 95.31 & 97.92 & 97.52 & 96.92 \\
 & 8 & 110 & 90.00 & 94.06 & 94.01 & 92.69 \\
 & 48 & 1 & 12.40 & 21.46 & 44.54 & 26.13 \\
 & 48 & 10 & 92.81 & 95.21 & 94.88 & 94.30 \\
 & 48 & 110 & 91.88 & 94.79 & 94.50 & 93.72 \\
 & Full & 1 & 11.56 & 18.85 & 42.21 & 24.21 \\
 & Full & 10 & 91.67 & 94.27 & 94.28 & 93.41 \\
 & Full & 110 & 91.56 & 93.96 & 94.50 & 93.34 \\
\bottomrule
\end{tabular}
\end{table}

\paragraph{Query-count interaction.}
For average accuracy $a(N,M)$ in percentage points,
\begin{equation}
  I =
  \bigl[a(\mathrm{Full},110)-a(8,110)\bigr]
  -
  \bigl[a(\mathrm{Full},1)-a(8,1)\bigr]
  = 4.07\ \text{points}.
  \label{eq:interaction}
\end{equation}
The $95\%$ paired question-bootstrap interval is $[2.00,6.28]$.
The per-benchmark interactions are $+4.79$, $+1.88$, and $+5.53$ points
on AIME24, AIME25, and AMC23. The pass@32 interaction is $6.45$ points
with interval $[0.71,12.69]$. These intervals quantify evaluation uncertainty
conditional on the trained models (Appendix~\ref{app:measurement:uncertainty}).

\subsection{Matched rollout sources at a direct 16K generation limit}
\label{app:matched-source-details}

Table~\ref{tab:refresh} tests the short-budget ordering in
Section~\ref{sec:behavior} using direct 16K generations on three benchmarks.
Current-policy sampling has higher mean accuracy under both teachers.
These samples are separate from the eight-benchmark 32K generations used
for the main budget curves.

\begin{table}[!htbp]
\centering
\caption{\textbf{Independent direct-16K evaluation on three benchmarks} (\%).
Current-policy mean accuracy exceeds initial-policy accuracy by 2.63 points
with Qwen3 (95\% interval $[-0.03,5.33]$) and 6.72 with JustRL
($[3.65,9.80]$).}
\label{tab:refresh}
\small
\setlength{\tabcolsep}{2.5pt}
\begin{tabular*}{\textwidth}{@{\extracolsep{\fill}}l*{3}{rr}rrr@{}}
\toprule
 & \multicolumn{2}{c}{AIME24} & \multicolumn{2}{c}{AIME25} & \multicolumn{2}{c}{AMC23} & \multicolumn{3}{c}{Average} \\
\cmidrule(lr){2-3}\cmidrule(lr){4-5}\cmidrule(lr){6-7}\cmidrule(lr){8-10}
Rollout source & Acc. & pass@32 & Acc. & pass@32 & Acc. & pass@32 & Acc. & pass@32 & Extraction \\
\midrule
\multicolumn{10}{l}{\textit{Teacher: Qwen3-4B-Instruct-2507}} \\
Current-policy & 11.35 & 50.00 & 15.83 & 40.00 & 43.49 & 81.93 & 23.56 & 57.31 & 66.72 \\
Initial-policy & 8.44 & 40.00 & 17.08 & 30.00 & 37.27 & 63.86 & 20.93 & 44.62 & 26.57 \\
\midrule
\multicolumn{10}{l}{\textit{Teacher: JustRL-DeepSeek-1.5B}} \\
Current-policy & 15.31 & 53.33 & 19.48 & 46.67 & 50.64 & 84.34 & 28.48 & 61.45 & 63.41 \\
Initial-policy & 10.00 & 33.33 & 16.67 & 26.67 & 38.59 & 56.63 & 21.75 & 38.88 & 24.16 \\
\bottomrule
\end{tabular*}
\end{table}

\subsection{Accuracy and generation behavior across response-length budgets}
\label{app:budget-curves}

The same trained models are evaluated on eight benchmarks to 32K, with shorter
budgets scored from saved-response prefixes. Table~\ref{tab:datasetdelta}
expands Figure~\ref{fig:budgetcrossover}a by benchmark and reports the
initial-policy/current-policy mean-length ratios at 32K. Benchmark weighting and
sample counts follow Appendix~\ref{app:measurement:suites}.

\begin{figure}[!htbp]
  \centering
  \includegraphics[width=\textwidth]{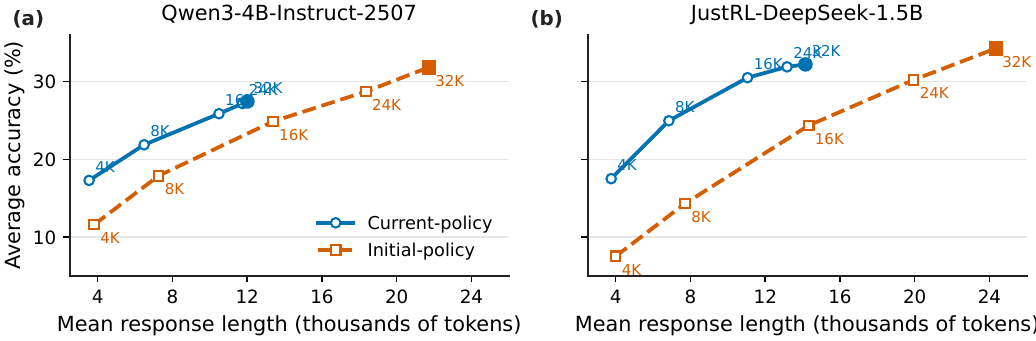}
  \caption{\textbf{Accuracy and response length across generation budgets.}
  Each curve plots eight-benchmark mean accuracy against mean response length
  at 4K, 8K, 16K, 24K, and 32K; filled markers denote 32K. Color and line
  style identify the rollout source, and the panels separate the two teachers.
  The horizontal axis uses decimal thousands of tokens. Per-benchmark accuracy
  differences and 32K length ratios appear in Table~\ref{tab:datasetdelta};
  all aggregates weight benchmarks equally.}
  \label{fig:tokeneff}
\end{figure}

At 16K, initial-policy/current-policy mean-length ratios are 1.28 and
1.30 under Qwen3 and JustRL; at 32K they increase to 1.81 and 1.72.
Figure~\ref{fig:tokeneff} plots accuracy against the actual mean retained
length. Initial-policy training reaches its higher final accuracy with
longer responses, whereas current-policy accuracy saturates earlier.
These are the rollout-source experiments of Section~\ref{sec:behavior}.

\begin{table}[H]
  \centering
  \caption{\textbf{Accuracy differences and 32K response-length ratios by benchmark.} Accuracy differences are initial-policy minus current-policy in percentage points. Positive values favor initial-policy training. Ratios are initial-policy/current-policy mean response length at 32K. Scores use prefixes of saved 32K responses; sample counts and benchmark weighting follow Appendix~\ref{app:measurement:suites}.}
  \label{tab:datasetdelta}
  \small
  \setlength{\tabcolsep}{4pt}
  \begin{tabular}{llrrrrrr}
    \toprule
    Teacher & Benchmark & 4K & 8K & 16K & 24K & 32K & Length ratio at 32K \\
    \midrule
    \multirow{8}{*}{Qwen3-4B} & AIME24 & -4.06 & -3.23 & -1.56 & +2.40 & +6.77 & $1.76\times$ \\
     & AIME25 & -4.06 & -3.85 & +1.25 & +3.65 & +5.94 & $1.82\times$ \\
     & AIME26 & -3.12 & -2.60 & +0.94 & +1.04 & +2.71 & $1.67\times$ \\
     & AMC23 & -11.11 & -6.36 & -4.48 & -0.94 & +4.33 & $1.85\times$ \\
     & HMMT25-Nov & -1.25 & -1.15 & +0.10 & +2.50 & +4.06 & $1.79\times$ \\
     & HMMT26-Feb & -2.75 & -3.22 & -2.37 & -0.28 & +2.08 & $1.83\times$ \\
     & MATH-500 & -10.40 & -3.80 & +0.65 & +2.30 & +5.05 & $2.06\times$ \\
     & OlympiadBench & -8.74 & -7.74 & -2.48 & +1.59 & +4.19 & $1.95\times$ \\
    \midrule
    \multirow{8}{*}{JustRL-1.5B} & AIME24 & -3.85 & -8.44 & -6.77 & -1.25 & +6.04 & $1.57\times$ \\
     & AIME25 & -7.50 & -11.35 & -4.48 & -0.94 & +1.67 & $1.73\times$ \\
     & AIME26 & -4.06 & -8.96 & -3.75 & +0.21 & +2.40 & $1.65\times$ \\
     & AMC23 & -22.06 & -17.43 & -12.42 & -6.81 & -0.75 & $1.84\times$ \\
     & HMMT25-Nov & -0.62 & -2.60 & -3.33 & -0.10 & +2.71 & $1.59\times$ \\
     & HMMT26-Feb & -1.61 & -6.91 & -3.98 & -1.89 & +0.00 & $1.74\times$ \\
     & MATH-500 & -24.35 & -15.40 & -5.25 & +0.55 & +3.35 & $2.12\times$ \\
     & OlympiadBench & -15.85 & -14.07 & -9.22 & -3.22 & +1.11 & $1.94\times$ \\
    \bottomrule
  \end{tabular}
\end{table}

\FloatBarrier

\section{State Sources, Order, and Additional Controls}
\label{app:replay}

Sections~\ref{sec:initial-states} and~\ref{sec:replay} ask whether early
states retain useful supervision, which source to replay, and in what
order. We expand the fixed-prefix measurements and organize the controls
by these questions.

\subsection{Shared continuation protocol}

All experiments in this appendix use the same 48 queries, Qwen3-0.6B
student, Qwen3-4B-Instruct-2507 teacher, and 16K training response limit.
The objective and optimizer settings follow Appendix~\ref{app:protocol}.
Each continuation restores both the model and AdamW state. A 20-step
stage consumes 2,400 responses in five 480-response iterations, each
containing optimizer minibatches of 128, 128, 128, and 96 responses.
Student candidates, probabilities, teacher scores, and loss coefficients
are recomputed before every scoring batch.

The initial-policy continuation bank contains ten responses per query in
each of five batches. It is collected separately from the 4,800 responses
used by the original 40-step initial-policy run. Continuations from both
40-step histories and the early replay experiment share this same
2,400-response bank. The fixed recent bank uses the same allocation but
is generated by $R_{40}$, the student after 40 current-policy steps.

Evaluation uses the eight-benchmark suite in
Appendix~\ref{app:measurement:suites}, with one training run and one
32K evaluation response set per condition. All shorter budgets re-score
prefixes of that set. Metric definitions and equal-benchmark averaging
follow Appendix~\ref{app:measurement:metrics}.

\subsection{Fixed-prefix diagnostics}
\label{app:replay:cross-scoring}

Diagnostic responses are sampled independently of the training banks.
For each of $\theta_0$, $R_{20}$, and $R_{40}$, we collect eight
responses per query and sample 32 positions uniformly without
replacement per response. The resulting 384 responses and 12,288
positions per source are held fixed across scoring models.
Full-vocabulary student and teacher probabilities are evaluated in FP32.
For scoring model $i$ and generating policy $j$, define
\begin{equation}
  E_{i,j}=\mathbb E_{s\sim d_j}
  \left[\DKL\!\left(\pi_i(\cdot\mid s)\,\Vert\,
  \pi_T(\cdot\mid s)\right)\right],
  \label{eq:rf-cross-kl}
\end{equation}
where $d_j$ weights responses equally and then sampled positions equally
within each response. This diagnostic differs from the top-16 training
objective. Table~\ref{tab:rf-cross-kl} reports the complete matrix;
continuation conditions are defined in Tables~\ref{tab:rf-switch}
and~\ref{tab:rf-state}.
The matrix shows residual disagreement on initial prefixes after current-policy
training and a larger reduction there when those prefixes are replayed.

\begin{table}[!htbp]
  \centering
  \caption{\textbf{Complete KL matrix on fixed diagnostic prefixes.}
  Full-vocabulary reverse KL (nats; lower is better), $E_{i,j}$ in
  Equation~\ref{eq:rf-cross-kl}, for the Qwen3-0.6B student and fixed
  Qwen3-4B-Instruct-2507 teacher. Rows specify the student used for scoring;
  columns specify the student that generated the diagnostic responses.
  Each column uses the same 384 responses and sampled positions across all
  scoring students, averaging positions within each response and then
  weighting responses equally. Steps denote total optimization steps; Start identifies
  the checkpoint restored for each 20-step continuation.}
  \label{tab:rf-cross-kl}
  \small
  \setlength{\tabcolsep}{4pt}
  \begin{tabular}{@{}llrrrr@{}}
    \toprule
    Start & Scoring student / continuation & Steps & Initial & $R_{20}$ & $R_{40}$ \\
    \midrule
    \multirow{3}{*}{---} & Initial student & 0 & 0.774 & 0.624 & 0.628 \\
     & $R_{20}$ & 20 & 0.548 & 0.505 & 0.523 \\
     & $R_{40}$ & 40 & 0.448 & 0.409 & 0.427 \\
    \midrule
    \multirow{3}{*}{$R_{40}$} & Continue current-policy sampling & 60 & 0.404 & 0.361 & 0.370 \\
     & Replay fixed recent rollouts & 60 & 0.404 & 0.360 & 0.370 \\
     & Replay initial-policy rollouts & 60 & 0.370 & 0.378 & 0.400 \\
    \midrule
    \multirow{1}{*}{$R_{20}$} & Replay initial-policy rollouts & 40 & 0.410 & 0.440 & 0.472 \\
    \bottomrule
  \end{tabular}
\end{table}

For checkpoints $a$ and $b$, the change in KL on their own responses is
\begin{equation}
 E_{a,a}-E_{b,b}
 =\underbrace{E_{a,a}-E_{b,a}}_{\text{fitting on fixed states}}
 +\underbrace{E_{b,a}-E_{b,b}}_{\text{change of state source}}.
 \label{eq:rf-fit-shift}
\end{equation}
The first term compares students on the same prefixes; the second changes
the prefix source while holding the scoring student fixed. For
$\theta_0\!\to R_{20}$, the total decrease is $0.268937$ nats, decomposed
into $0.226405$ from fixed-state fitting and $0.042532$ from the state-source
change. For $R_{20}\!\to R_{40}$, the corresponding values are $0.078024$,
$0.096221$, and $-0.018198$ nats.

\paragraph{Concentration of the KL reduction.}
We rank the 12,288 initial-policy positions by their KL under $R_{40}$.
On the highest 1,229 positions, replay after 40 steps and continued
current-policy training have mean KL 2.437975 and 2.758502 nats;
on the remaining 11,059, they have 0.140661 and 0.142670 nats.
The upper group contributes 94.66\% of the signed sum of KL differences
(continued current-policy training minus replay). Replay has lower KL
on 41.57\% of individual positions and 77.60\% of response-level averages.
This top-10\% partition is a post-hoc description of the net difference.

\paragraph{Exact-prefix likelihood.}
Prefix probability is the product of token probabilities along the
recorded sequence. Among the 12,260 nonempty sampled initial-policy
prefixes, probability is lower than at initialization for 99.15\%
under $R_{40}$ and 99.29\% after its 20-step initial-policy continuation,
using equal response weights. These measurements concern exact recorded
prefixes; they do not measure the probability of semantically equivalent
responses.

\subsection{Continuation source after two training histories}

For the four history comparisons, the first 40 steps use either initial-policy or current-policy rollouts;
the final 20 use either the shared initial-policy bank or newly sampled
current-policy responses. Table~\ref{tab:rf-switch} lists these histories
and continuations explicitly. Comparisons within a starting history are
matched. Comparing opposite switches across histories changes the total
allocation of the two sources and is not the equal-allocation order test.

\begin{table}[!htbp]
  \centering
\caption{\textbf{Continuation source after 40 optimization steps.}
Eight-benchmark average accuracy (\%) across response budgets. All endpoints
have 60 steps; mean response length uses complete 32K responses.
Comparisons sharing a first-stage source restore the same model and optimizer.
The fixed recent bank is generated by the 40-step current-policy student.}
\label{tab:rf-switch}
\label{tab:rf-switch-budgets}
\label{tab:rf-switch-metrics}
\small
\setlength{\tabcolsep}{3pt}
\begin{tabular}{lrrrrrr}
    \toprule
    Training history & 4K & 8K & 16K & 24K & 32K & Tokens \\
    \midrule
    Initial $\rightarrow$ initial & 12.85 & 18.66 & 25.24 & 28.93 & 30.31 & 21,357 \\
    Initial $\rightarrow$ current & 17.67 & 22.87 & 26.62 & 27.58 & 27.71 & 11,166 \\
    Current $\rightarrow$ initial & 14.23 & 20.71 & 27.13 & 30.06 & 29.37 & 12,116 \\
    Current $\rightarrow$ current & 17.94 & 23.06 & 26.41 & 26.52 & 26.52 & 8,254 \\
    Current $\rightarrow$ fixed recent & 18.15 & 23.10 & 26.12 & 26.20 & 26.21 & 7,997 \\
    \bottomrule
  \end{tabular}
\end{table}

\begin{table}[!htbp]
  \centering
  \caption{\textbf{Per-benchmark accuracy for the four continuations (\%).} Each column lists the first 40 and final 20 steps, as defined in Table~\ref{tab:rf-switch}. Average weights benchmarks equally and uses unrounded values.}
  \label{tab:rf-switch-datasets}
  \small
  \setlength{\tabcolsep}{4pt}
  \begin{tabular}{lrrrr}
    \toprule
    Benchmark & \shortstack{Initial then\\initial} & \shortstack{Initial then\\current} & \shortstack{Current then\\initial} & \shortstack{Current then\\current} \\
    \midrule
    \multicolumn{5}{c}{16K retained prefixes} \\
    \midrule
    AIME24 & 10.10 & 11.04 & 11.15 & 12.71 \\
    AIME25 & 16.67 & 16.25 & 16.15 & 16.77 \\
    AIME26 & 13.23 & 12.92 & 13.02 & 10.83 \\
    AMC23 & 38.48 & 42.73 & 43.83 & 44.69 \\
    HMMT26-Feb & 5.68 & 9.09 & 8.71 & 8.43 \\
    HMMT25-Nov & 6.04 & 6.77 & 6.67 & 5.73 \\
    MATH-500 & 72.35 & 73.40 & 75.85 & 71.85 \\
    OlympiadBench & 39.41 & 40.74 & 41.67 & 40.26 \\
    \midrule
    Average & 25.25 & 26.62 & 27.13 & 26.41 \\
    \midrule
    \multicolumn{5}{c}{32K generations} \\
    \midrule
    AIME24 & 17.08 & 12.71 & 15.10 & 12.71 \\
    AIME25 & 19.79 & 16.98 & 16.88 & 16.88 \\
    AIME26 & 14.69 & 13.85 & 14.48 & 10.94 \\
    AMC23 & 50.15 & 44.31 & 48.61 & 44.88 \\
    HMMT26-Feb & 9.66 & 10.32 & 10.98 & 8.43 \\
    HMMT25-Nov & 8.75 & 7.71 & 6.77 & 5.83 \\
    MATH-500 & 77.20 & 73.85 & 77.85 & 71.95 \\
    OlympiadBench & 45.19 & 41.93 & 44.30 & 40.56 \\
    \midrule
    Average & 30.31 & 27.71 & 29.37 & 26.52 \\
    \bottomrule
  \end{tabular}
\end{table}

The direction of the 16K continuation effect differs between the two
starting histories. At 32K, initial-policy continuation is more accurate
from both histories. After current-policy training, replay improves
accuracy on five of eight benchmarks at 16K and seven at 32K,
with a tie on AIME25 at 32K.

\subsection{Early replay at a matched budget}
\label{app:replay:stage}

Table~\ref{tab:rf-state} complements the late-replay comparison in
Section~\ref{sec:replay:source}. Early replay adds 20 initial-policy steps
to $R_{20}$; its current-policy control is $R_{40}$. Both endpoints use
40 updates and 4,800 trajectories. Early replay improves 32K accuracy
but lowers 16K accuracy, while late replay improves both budgets relative
to its own current-policy control (Table~\ref{tab:rf-switch}). The early
replay endpoint also supplies the full-objective baseline for the EOS
controls below.

\begin{table}[!htbp]
  \centering
\caption{\textbf{Early replay at a matched 40-update budget.}
Eight-benchmark average accuracy (\%) across response budgets. Current only is
$R_{40}$; early replay adds 20 initial-policy steps to $R_{20}$. Mean response length
uses complete 32K responses.}
\label{tab:rf-state}
\label{tab:rf-state-budgets}
\small
\setlength{\tabcolsep}{3pt}
\begin{tabular}{lrrrrrrr}
    \toprule
    Condition & Steps & 4K & 8K & 16K & 24K & 32K & Tokens \\
    \midrule
    Current only & 40 & 17.28 & 21.86 & 25.86 & 27.18 & 27.44 & 12,011 \\
    Replay after 20 steps & 40 & 11.20 & 17.10 & 24.40 & 28.89 & 32.02 & 22,343 \\
    \bottomrule
  \end{tabular}
\end{table}

\subsection{Training order at fixed source allocation}
\label{app:replay:controls}

The order comparison in Section~\ref{sec:replay:order} restores the same
$R_{20}$ model and optimizer, then applies two 20-step stages in opposite
orders. Table~\ref{tab:state-source-datasets} expands the main-table
comparison by benchmark.

\paragraph{Order at fixed allocation.}
Both orders contain 40 current-policy and 20 initial-policy steps in
total. Current-policy responses are generated at their respective
stages, so changing the order also changes those responses. Current then
initial is more accurate on all eight benchmarks at 16K, 24K, and 32K,
with mean gains of 1.61 and 3.31 points at 16K and 32K
(Table~\ref{tab:state-source-datasets}).
The reverse order is stronger on all eight benchmarks at 4K and seven at
8K. The order-control and continuation-source endpoints use distinct
optimizer histories and are evaluated separately.

\begin{table}[!htbp]
  \centering
  \caption{\textbf{Per-benchmark advantage of current-policy training followed
  by initial-policy replay.} Values are accuracy differences in percentage
  points relative to the reverse order; positive values favor current then
  initial. The average weights benchmarks equally.}
  \label{tab:state-source-datasets}
  \small
  \begin{tabular}{lrrrr}
    \toprule
    Benchmark & 4K & 8K & 16K & 32K \\
    \midrule
    AIME24 & -1.67 & -1.77 & +1.46 & +4.90 \\
    AIME25 & -4.69 & -2.29 & +1.04 & +0.94 \\
    AIME26 & -3.54 & -1.67 & +2.40 & +3.33 \\
    AMC23 & -6.40 & -3.84 & +2.52 & +6.59 \\
    HMMT25-Nov & -0.31 & +1.04 & +1.46 & +2.29 \\
    HMMT26-Feb & -1.42 & -1.04 & +0.76 & +2.27 \\
    MATH-500 & -5.95 & -1.00 & +2.50 & +3.65 \\
    OlympiadBench & -4.67 & -3.78 & +0.74 & +2.48 \\
    \midrule
    Average & -3.58 & -1.79 & +1.61 & +3.31 \\
    \bottomrule
  \end{tabular}
\end{table}
\FloatBarrier

\subsection{Mixed sources and direct EOS supervision}

These controls extend the source comparison in Section~\ref{sec:replay:source}:
they test how loss weighting affects mixed-source learning and whether the
direct EOS term accounts for early replay's accuracy and length changes.
All conditions in Table~\ref{tab:state-source-key} restore $R_{20}$ and add
20 updates using the same initial-policy bank.

\begin{table}[!htbp]
\centering\small
\caption{\textbf{Loss controls for 20 replay-stage updates after $R_{20}$.}
Eight-benchmark accuracy (\%) and mean 32K output length. All endpoints
use 40 updates in total. Full replay is the early-replay baseline in
Table~\ref{tab:rf-state}; the mixtures use equal response counts per source.}
\label{tab:state-source-key}
\label{tab:state-source-budgets}
\label{tab:state-source-metrics}
\begin{tabular}{lrrr}
\toprule
Continuation & 16K & 32K & Tokens at 32K \\
\midrule
Full initial-policy replay & 24.40 & 32.02 & 22,343 \\
\midrule
Equal-response mix & 26.38 & 28.25 & 12,346 \\
Source-balanced mix & 27.00 & 30.09 & 14,445 \\
\midrule
Replay: no EOS & 25.07 & 31.88 & 22,111 \\
Replay: EOS only & 23.64 & 25.57 & 13,870 \\
\bottomrule
\end{tabular}
\end{table}

\paragraph{Mixed-source loss reduction.}
Both mixtures draw equal response counts from the two sources.
The equal-response condition uses the original dynamic-microbatch
reduction. The source-balanced condition computes an all-rank token mean
separately for each source in each optimizer minibatch and weights each
by $0.5$. It retains the final minibatch's $96/128=0.75$ scaling.
Source balancing increases accuracy and response length at 32K while
reducing answer extraction from 97.50\% to 96.33\%.

\paragraph{Direct EOS-candidate masks.}
The EOS controls retain the initial-policy text and original student-top-16
candidate sets. ``No EOS'' sets the coefficient of EOS (token 151645)
to zero where it already occurs among the candidates; ``EOS only''
retains only that coefficient. Neither inserts EOS into a candidate set.
Their matched full-objective baseline is early replay after $R_{20}$
(Table~\ref{tab:rf-state}). Removing the direct EOS term leaves 32K
accuracy and response length close to full replay. Retaining only that
term yields shorter responses and higher answer extraction, but lower
accuracy. These masks isolate direct loss terms; shared parameters and
softmax normalization still couple their effects on predictions.

\subsection{Complementary gradient signals on earlier states}
\label{sec:mechanism}

As a secondary diagnostic for the source effects in Section~\ref{sec:replay},
we hold the full-query, per-step-resampling model fixed at updates 0, 55,
and 110. We score 128 trajectories from 48 queries in 16 minibatches of
eight. Current-policy responses come from the checkpoint being probed;
initial-policy responses come from $\theta_0$. Candidate sets, probabilities,
and gradients are recomputed without an optimizer update. Norms use all
parameters before clipping and AdamW; cosines use the same 3,712 coordinates
(16 per parameter tensor and FSDP rank) and pair minibatches by index.

\begin{table}[!htbp]
  \centering
  \caption{\textbf{Gradients on fixed model parameters.} Means and standard
  deviations are over 16 diagnostic minibatches. Norms use all parameters;
  cross-source cosine uses a fixed subset of gradient coordinates. Parentheses
  give its standard error over paired minibatches.}
  \label{tab:gradients}
  \small
  \begin{tabular}{lrr}
    \toprule
    Condition & exact $\|g\|_2$ & \shortstack{Cosine vs.\\current-policy} \\
    \midrule
    $\theta_0$ current-policy        & $19.426 \pm 8.018$ & --- \\
    $\theta_{55}$ current-policy     & $5.419 \pm 2.525$  & --- \\
    $\theta_{55}$ initial-policy     & $13.717 \pm 4.781$ & $0.127\ (0.092)$ \\
    $\theta_{110}$ current-policy    & $5.061 \pm 3.090$  & --- \\
    $\theta_{110}$ initial-policy    & $14.261 \pm 5.043$ & $0.034\ (0.078)$ \\
    \bottomrule
  \end{tabular}
\end{table}

At updates 55 and 110, initial-policy gradient norms are 2.53 and 2.82
times their current-policy counterparts, with near-zero paired cosines.
Thus the two sources provide different update signals at the same model
parameters, complementing the fixed-prefix KL and continuation results.
The online trigger in Section~\ref{sec:annealing:trigger} instead compares
successive current-policy iterations while parameters are updated.

\FloatBarrier

\section{\ropd{}: Implementation and Complete Results}
\label{app:annealing}

This appendix makes the curriculum in Section~\ref{sec:annealing}
reproducible and expands its evidence: paired runs support the main means
and trigger statistics, while individual-run tables resolve budget and
benchmark variation.

\subsection{Source distribution and its interpretation}
\label{sec:annealing:source}
\label{app:annealing:algebra}

Let $d_{\theta_k}$ be the state distribution generated by the current
student and $d_{\mathrm{init}}$ the fixed empirical distribution of the
initial-policy rollout bank. With $\alpha_k$ defined by
Equation~\ref{eq:anneal-schedule}, the training source is
\begin{equation}
\mu_k=
\begin{cases}
d_{\theta_k}, & k\leq\tau,\\
d_{\mathrm{init}}, & \tau<k\leq K.
\end{cases}
\label{eq:state-mixture}
\end{equation}
Each iteration uses one source for all four optimizer minibatches.
After the switch, response text comes entirely from the initial-policy
bank; student probabilities, teacher scores, and distillation coefficients
are recomputed. If no trigger occurs, all iterations use current-policy
rollouts.

\subsection{Training algorithm and gradient measurement}
\label{app:annealing:implementation}

\begin{algorithm}[htbp]
 \caption{\ropd{}}
 \label{alg:annealing}
 \centering
 \begin{minipage}{0.96\textwidth}
 \begin{enumerate}
  \setlength{\itemsep}{2pt}
  \item \textbf{Initialize:} Set the student to $\theta_0$, save 2,400
  initial-policy responses, and leave $\tau$ unset.
  \item \textbf{Select source ($k=1,\ldots,K$):} Generate 480 current-policy rollouts if the
  trigger is unset. Otherwise, take 480 rollouts from the initial-policy
  bank. Use this source for all four optimizer minibatches.
  \item \textbf{Train:} Recompute student candidates and student/teacher
  probabilities. Apply four AdamW updates using the shared distillation
  loss; while $\tau$ is unset, record each gradient before clipping and AdamW.
  \item \textbf{Check trigger:} While $\tau$ is unset, compute
  $(\bar g_k,v_k,D_k,V_k)$. Two consecutive comparisons satisfying
  Equation~\ref{eq:anneal-trigger} set $\tau=k$; replay starts at $k+1$.
  \item \textbf{Finish:} Repeat steps 2--4 through iteration $K$ and evaluate
  the final checkpoint.
 \end{enumerate}
 \end{minipage}
\end{algorithm}

The trigger compares consecutive available current-policy gradient
measurements using $D_k\leq V_k$ and requires two consecutive qualifying
comparisons before replay begins, when batches contain only current-policy
rollouts. Before a comparison is available, training uses current-policy rollouts. If
$V_k=0$, only $D_k=0$ qualifies. An unset trigger keeps training on current-policy
rollouts; a trigger at the final iteration leaves no replay iterations. A recorded
trigger is not reversed by subsequent ratios above one.

Each 480-response iteration contains optimizer minibatches of
128, 128, 128, and 96 responses. Gradients are recorded before clipping
and AdamW. The trainer scales the final minibatch's token-mean loss by
$96/128$; the diagnostic copy divides out this factor for comparison,
while the actual optimizer gradient retains it. Running means and sums
of squared deviations are accumulated over local FP32 parameter-gradient
shards; scalar reductions across eight ranks recover full-model norms
and squared differences.

The four measured gradients include both minibatch variation and
parameter changes caused by earlier updates within the iteration.
Thus $V_k$ is an empirical comparison scale, unlike the fixed-parameter,
selected-coordinate diagnostic in Appendix~\ref{sec:mechanism}.

\begin{table}[htbp]
 \centering
 \small
 \caption{\textbf{Curriculum settings shared across student sizes.}
 Each student generates its own initial-policy bank and measures its own gradients.}
 \label{tab:annealing-config}
 \begin{tabular}{ll}
 \toprule
 Quantity & Value \\
 \midrule
 0.6B / 1.7B teacher & Qwen3-4B-Instruct-2507 \\
 8B teacher & Qwen3-30B-A3B-Instruct-2507 \\
 Query set & 48 random queries; ten occurrences each per iteration \\
 Iterations / optimization steps & 15 / 60 \\
 Responses per iteration / initial bank & 480 / five batches of 480 \\
 Trigger threshold / persistence & $D\leq V$ / two consecutive comparisons \\
 Source after triggering & Initial-policy bank for every $k>\tau$ \\
 \bottomrule
 \end{tabular}
\end{table}

Optimizer settings follow Table~\ref{tab:setup-config}; evaluation follows
Appendix~\ref{app:measurement}. Each iteration uses 480 responses, totaling
7,200 trajectory uses.
A trigger at $\tau$ gives $480\tau$ current-policy trajectory uses and
$480(K-\tau)$ initial-policy replay uses; without a trigger, all uses
are current-policy. Compute includes initial-bank construction, current-policy
generation, teacher scoring, and student optimization.

\subsection{Paired data-order repetitions and trigger iterations}
\label{app:annealing:repeats}

Table~\ref{tab:annealing-seed-results} reports three paired runs
for each student size. Every row uses the same 48-query
set with a different training-data order; $\tau$ is the R-OPD trigger
iteration (four optimizer updates per iteration). The main table reports
the mean and sample SD of these accuracies. Its Tokens column averages
the mean 32K output lengths from the same three runs and displays whole
tokens. At 16K and 32K, R-OPD exceeds fixed
replay in each of the three runs at every size. The trigger sequences are
$(5,5,7)$, $(7,6,7)$, and $(7,8,8)$ for 0.6B, 1.7B, and 8B.

\begin{table}[!htbp]
\centering\small
\setlength{\tabcolsep}{4pt}
\caption{\textbf{Paired accuracy and R-OPD trigger by training-data order.}
Eight-benchmark average accuracy (\%); each row gives one training-data
shuffle of the same 48 queries. $\tau$ is in training iterations.}
\label{tab:annealing-seed-results}
\begin{tabular}{llrrrrrrr}
\toprule
Student & Run & $\tau$ & \multicolumn{3}{c}{16K accuracy (\%)} & \multicolumn{3}{c}{32K accuracy (\%)} \\
\cmidrule(lr){4-6}\cmidrule(l){7-9}
 & & & Current & Fixed & \ropd{} & Current & Fixed & \ropd{} \\
\midrule
0.6B & 1 & 5 & 26.41 & 27.13 & 28.96 & 26.52 & 29.37 & 31.52 \\
 & 2 & 5 & 25.94 & 27.52 & 28.38 & 26.07 & 29.62 & 30.48 \\
 & 3 & 7 & 26.77 & 27.49 & 28.53 & 26.91 & 29.68 & 30.83 \\
\midrule
1.7B & 1 & 7 & 38.97 & 41.21 & 41.92 & 39.13 & 43.74 & 45.14 \\
 & 2 & 6 & 38.73 & 41.27 & 41.65 & 38.86 & 43.62 & 45.43 \\
 & 3 & 7 & 39.51 & 41.10 & 42.06 & 39.55 & 43.65 & 45.42 \\
\midrule
8B & 1 & 7 & 60.37 & 60.07 & 60.65 & 61.63 & 63.60 & 65.37 \\
 & 2 & 8 & 60.46 & 60.59 & 61.42 & 61.60 & 64.06 & 65.77 \\
 & 3 & 8 & 60.13 & 60.04 & 61.24 & 61.48 & 63.66 & 65.72 \\
\bottomrule
\end{tabular}

\end{table}

\subsection{Fixed replay-onset scan for Qwen3-0.6B}
\label{app:annealing:fixed-onset}

Table~\ref{tab:annealing-fixed-onset}b varies the fixed switch iteration
within the same 15-iteration training configuration. A switch at $\tau$
uses current-policy rollouts through that iteration and the same
initial-policy bank thereafter; $\tau=0$ and $\tau=15$ are the
initial-only and current-only endpoints. Each schedule averages the same
three training-data orders; the final column reports \ropd{}.

\subsection{Individual-run budgets and benchmark variation}
\label{app:annealing:results}
\label{app:annealing:transfer}
\label{app:annealing:8b}

Tables~\ref{tab:annealing-datasets} and~\ref{tab:annealing-original-datasets}
expand individual endpoints for each student size;
Table~\ref{tab:annealing-seed-results} provides the paired repetitions. Each endpoint uses 60 updates and
7,200 trajectory uses. Shorter budgets re-score saved 32K responses;
Tokens reports that individual endpoint's mean output length. The main
Table~\ref{tab:annealing-main} instead aggregates three data-order runs.
The 0.6B fixed continuation is the current-to-initial source control in
Table~\ref{tab:rf-switch}; the separately resumed order comparison remains
in Appendix~\ref{app:replay:controls}.

\begin{table}[!htbp]
\centering\small
\setlength{\tabcolsep}{4pt}
\caption{\textbf{Individual-run accuracy across response budgets.}
Eight-benchmark mean accuracy (\%) and mean 32K output length for one
trained endpoint per method and student size. Fixed denotes $R_{40}F_{20}$.
Teachers: Qwen3-4B-Instruct-2507 for 0.6B/1.7B;
Qwen3-30B-A3B-Instruct-2507 for 8B.}
\label{tab:annealing-datasets}
\label{tab:annealing-budgets}
\label{tab:annealing-transfer-datasets}
\label{tab:annealing-transfer-budgets}
\label{tab:annealing-8b-budgets}
\begin{tabular}{llrrrrrr}
\toprule
Student & Method & 4K & 8K & 16K & 24K & 32K & Tokens at 32K \\
\midrule
0.6B & Current & 17.94 & 23.06 & 26.41 & 26.52 & 26.52 & 8,254 \\
 & Fixed & 14.23 & 20.71 & 27.13 & 30.06 & 29.37 & 12,116 \\
 & \ropd{} & 14.72 & 21.61 & 28.96 & 31.28 & 31.52 & 13,842 \\
\midrule
1.7B & Current & 29.06 & 35.59 & 38.97 & 39.13 & 39.13 & 6,887 \\
 & Fixed & 27.11 & 34.52 & 41.21 & 44.44 & 43.74 & 11,025 \\
 & \ropd{} & 26.77 & 34.66 & 41.92 & 44.61 & 45.14 & 11,980 \\
\midrule
8B & Current & 38.62 & 51.80 & 60.37 & 61.58 & 61.63 & 7,793 \\
 & Fixed & 36.65 & 48.45 & 60.07 & 63.98 & 63.60 & 10,363 \\
 & \ropd{} & 36.95 & 49.28 & 60.65 & 64.79 & 65.37 & 11,094 \\
\bottomrule
\end{tabular}
\end{table}

The budget table resolves the crossover shown in Figure~\ref{fig:curriculum-profile}:
the current-policy advantage at short prefixes gives way to replay gains
at longer budgets. The benchmark breakdown shows that the 8B mean gain
includes both improvements and small reversals on individual benchmarks.

\begin{table}[!htbp]
\centering\small
\setlength{\tabcolsep}{4pt}
\caption{\textbf{Benchmark variation in the individual runs.}
Accuracy (\%) for the endpoints in Table~\ref{tab:annealing-datasets}.
Fixed denotes $R_{40}F_{20}$. Average weights the eight benchmarks equally.}
\label{tab:annealing-original-datasets}
\label{tab:annealing-8b-datasets}
\begin{tabular}{llrrrrrr}
\toprule
Student & Benchmark & \multicolumn{3}{c}{16K} & \multicolumn{3}{c}{32K} \\
\cmidrule(lr){3-5}\cmidrule(l){6-8}
 & & Current & Fixed & \ropd{} & Current & Fixed & \ropd{} \\
\midrule
0.6B & AIME24 & 12.71 & 11.15 & 13.33 & 12.71 & 15.10 & 16.56 \\
 & AIME25 & 16.77 & 16.15 & 18.44 & 16.88 & 16.88 & 19.90 \\
 & AIME26 & 10.83 & 13.02 & 15.31 & 10.94 & 14.48 & 17.29 \\
 & AMC23 & 44.69 & 43.83 & 46.69 & 44.88 & 48.61 & 51.77 \\
 & HMMT25-Nov & 5.73 & 6.67 & 7.40 & 5.83 & 6.77 & 8.85 \\
 & HMMT26-Feb & 8.43 & 8.71 & 10.61 & 8.43 & 10.98 & 12.88 \\
 & MATH-500 & 71.85 & 75.85 & 76.80 & 71.95 & 77.85 & 78.85 \\
 & OlympiadBench & 40.26 & 41.67 & 43.07 & 40.56 & 44.30 & 46.04 \\
\cmidrule(l){2-8}
 & Average & 26.41 & 27.13 & 28.96 & 26.52 & 29.37 & 31.52 \\
\midrule
1.7B & AIME24 & 33.75 & 36.77 & 37.92 & 33.96 & 41.25 & 42.60 \\
 & AIME25 & 24.06 & 27.40 & 26.77 & 24.17 & 28.96 & 29.17 \\
 & AIME26 & 20.31 & 25.21 & 25.73 & 20.42 & 26.98 & 28.44 \\
 & AMC23 & 61.14 & 63.67 & 64.65 & 61.45 & 67.13 & 69.39 \\
 & HMMT25-Nov & 14.27 & 15.00 & 16.35 & 14.48 & 18.33 & 19.79 \\
 & HMMT26-Feb & 18.75 & 20.08 & 20.64 & 18.94 & 23.58 & 24.72 \\
 & MATH-500 & 85.95 & 86.30 & 87.55 & 85.95 & 87.15 & 89.05 \\
 & OlympiadBench & 53.56 & 55.30 & 55.78 & 53.70 & 56.56 & 58.00 \\
\cmidrule(l){2-8}
 & Average & 38.97 & 41.21 & 41.92 & 39.13 & 43.74 & 45.14 \\
\midrule
8B & AIME24 & 62.60 & 60.31 & 60.31 & 64.38 & 66.35 & 66.25 \\
 & AIME25 & 48.75 & 49.17 & 50.31 & 51.25 & 56.98 & 57.29 \\
 & AIME26 & 52.92 & 54.90 & 53.75 & 54.38 & 60.00 & 61.25 \\
 & AMC23 & 82.57 & 81.97 & 82.94 & 83.55 & 84.26 & 86.56 \\
 & HMMT25-Nov & 40.42 & 38.33 & 40.10 & 42.19 & 43.75 & 47.40 \\
 & HMMT26-Feb & 33.81 & 33.52 & 34.09 & 34.38 & 34.38 & 36.65 \\
 & MATH-500 & 94.30 & 94.80 & 95.10 & 94.60 & 94.40 & 95.95 \\
 & OlympiadBench & 67.59 & 67.59 & 68.56 & 68.37 & 68.70 & 71.59 \\
\cmidrule(l){2-8}
 & Average & 60.37 & 60.07 & 60.65 & 61.63 & 63.60 & 65.37 \\
\bottomrule
\end{tabular}
\end{table}

\FloatBarrier

\section{Extended Related Work}
\label{app:related}

OPD trains on student-visited states \citep{gu2024minillm,agarwal2024gkd}
and supports reasoning post-training \citep{yang2025qwen3,song2026survey}.
Recent analyses study divergence, candidate support, token corrections,
and teacher--student compatibility
\citep{armandpour2026unmasking,zhu2026manyfaces,fu2026failuremodes,
xie2026topd,wang2026teachability,li2026rethinkingopd}.
Methods that shorten responses, supply teacher prefixes, or select
supervision positions and samples
\citep{liu2026pwopsd,xie2026positionbias,zhou2026esr,
zhang2026prefixopd,xu2026relayopd}
change the supervision or the states receiving it. RL sample- and
token-selection methods \citep{wang2026d3s} address a related allocation
problem; our focus is rollout origin and training order.

Small-data reasoning studies show gains from compact training sets
\citep{ye2025limo,muennighoff2025s1,li2025limr,wang2025oneshotrlvr}.
In OPD, repeated responses to one query visit different prefixes;
query count, trajectories per query, and resampling frequency therefore
measure different aspects of training. This distinction motivates the
query-count and resampling grid in Section~\ref{sec:grid}.

Large-scale post-training systems also manage trajectory generation and
policy updates \citep{yu2025dapo,sheng2025verl}; Section~\ref{sec:related}
discusses the relation of our replay bank to rollout freshness and reuse.

\section{Limitations}
\label{sec:limitations}

Our experiments focus on competition mathematics with Qwen3 students.
Evaluating the same replay curriculum on other task families and student
architectures would establish how broadly these results transfer.

\end{document}